\documentclass[fleqn,10pt]{wlscirep}
\usepackage[utf8]{inputenc}
\usepackage[T1]{fontenc}

\usepackage{booktabs}
\usepackage{multirow}
\usepackage{soul}
\usepackage{bm}
\usepackage[normalem]{ulem}
\useunder{\uline}{\ul}{}
\usepackage{nameref}
\usepackage{hyperref}
\usepackage{caption}
\usepackage{subcaption}
\usepackage[para]{threeparttable}
\usepackage{pdflscape}
\usepackage{rotating}
\usepackage{tablefootnote}
\usepackage{textcomp}
\usepackage{amsmath}
\usepackage[edges]{forest}

\soulregister\cite7 

\definecolor{foldercolor}{RGB}{124,166,198}

\tikzset{pics/folder/.style={code={%
    \node[inner sep=0pt, minimum size=#1](-foldericon){};
    \node[folder style, inner sep=0pt, minimum width=0.3*#1, minimum height=0.6*#1, above right, xshift=0.05*#1] at (-foldericon.west){};
    \node[folder style, inner sep=0pt, minimum size=#1] at (-foldericon.center){};}
    },
    pics/folder/.default={20pt},
    folder style/.style={draw=foldercolor!80!black,top color=foldercolor!40,bottom color=foldercolor}
}

\forestset{is file/.style={edge path'/.expanded={%
        ([xshift=\forestregister{folder indent}]!u.parent anchor) |- (.child anchor)},
        inner sep=1pt},
    this folder size/.style={edge path'/.expanded={%
        ([xshift=\forestregister{folder indent}]!u.parent anchor) |- (.child anchor) pic[solid]{folder=#1}}, inner xsep=0.6*#1},
    folder tree indent/.style={before computing xy={l=#1}},
    folder icons/.style={folder, this folder size=#1, folder tree indent=3*#1},
    folder icons/.default={12pt},
}

\title{A dataset of high-resolution plantar pressures for gait analysis across varying footwear and walking speeds}

\author[1]{Robyn Larracy}
\author[1]{Angkoon Phinyomark}
\author[1]{Ala Salehi}
\author[1]{Eve MacDonald}
\author[1]{Saeed Kazemi}
\author[1]{Shikder Shafiul Bashar}
\author[1]{Aaron Tabor}
\author[1,*]{Erik Scheme}
\affil[1]{University of New Brunswick, Institute of Biomedical Engineering, Fredericton, E3B 5A3, Canada}

\affil[*]{corresponding author: Erik Scheme (escheme@unb.ca)}

\begin{abstract}
Gait refers to the patterns of limb movement generated during walking, which are unique to each individual due to both physical and behavioural traits. 
Walking patterns have been widely studied in biometrics, biomechanics, sports, and rehabilitation. 
While traditional methods rely on video and motion capture, advances in plantar pressure sensing technology now offer deeper insights into gait. 
However, underfoot pressures during walking remain underexplored due to the lack of large, publicly accessible datasets. 
To address this, we introduce the UNB StepUP-P150 dataset:
a footStep database for gait analysis and recognition using Underfoot Pressure, including data from 150 individuals.
This dataset comprises high-resolution plantar pressure data (4 sensors/cm\textsuperscript{2}) collected using a 1.2m by 3.6m pressure-sensing walkway. 
It contains over 200,000 footsteps from participants walking with various speeds (preferred, slow-to-stop, fast, and slow) and footwear conditions (barefoot, standard shoes, and two personal shoes),  supporting advancements in biometric gait recognition and presenting new research opportunities in biomechanics and deep learning. 
UNB StepUP-P150 establishes a new benchmark for plantar pressure-based gait analysis and recognition.
\end{abstract}

\begin{document}

\flushbottom
\maketitle

\thispagestyle{empty}

% \noindent Please note: Abbreviations should be introduced at the first mention in the main text – no abbreviations lists or tables should be included. Structure of the main text is provided below.

%%%%%%%%%%%%%%%%%%%%%%%%%%%%%%%%%%%%%%%%%%%%%%%%%%%%%%%%%%%%%%%%%%%%%%%%%%%%%%%%%%%
%%%%%%%%%%%%%%%%%%%%%%%%%%%%%%%%%%%%%%%%%%%%%%%%%%%%%%%%%%%%%%%%%%%%%%%%%%%%%%%%%%%
%%%%%%%%%%%%%%%%%%%%%%%%%%%%%%%%%%%%%%%%%%%%%%%%%%%%%%%%%%%%%%%%%%%%%%%%%%%%%%%%%%%
\section*{Background \& Summary}

%%% (700 words maximum) An overview of the study design, the assay(s) performed, and the created data, including any background information needed to put this study in the context of previous work and the literature. The section should also briefly outline the broader goals that motivated the creation of this dataset and the potential reuse value. We also encourage authors to include a figure that provides a schematic overview of the study and assay(s) design. The Background \& Summary should not include subheadings. This section and the other main body sections of the manuscript should include citations to the literature as needed. 

% Par 1: Gait Analysis
Gait analysis has long been a critical area of study within biometrics, biomechanics, sports, and rehabilitation. 
The unique patterns of human walking provide valuable insight into an individual's identity, health status, athletic performance, and even potential underlying medical conditions. 
Although traditional methods of gait analysis have relied heavily on video-based systems and motion capture technologies, recent advances in sensor technology have opened new avenues for capturing and analyzing gait patterns. 
Among these, plantar pressure measurement has appeared as a promising avenue due to its potential to provide detailed insights into footstep patterns, including the pressure exerted at different points under the foot and temporal changes in these patterns throughout the gait cycle. 
Despite its potential, this modality remains under-researched \cite{ConnorRoss2018}, especially with regard to the development of large, publicly available datasets, thereby limiting the exploration of modern advanced deep learning algorithms for gait analysis and recognition.

% Par 2: UNB StepUp database
Despite the availability of many other gait databases and datasets, the majority use video-based systems, motion capture technologies, and wearable devices \cite{Singh2018,Wan2018,SepasMoghaddamEtemad2023,dosSantos2023}. 
Databases that use floor sensor technology generally collect data from force plates, as seen in the GaitRec \cite{Horsak2020}, Gutenberg \cite{Horst2021}, and ForceID A \cite{Duncanson2023} datasets, which provide more constrained and less comprehensive information than the complex spatio-temporal data offered by emerging high-resolution pressure sensors. 
Conversely, databases that emphasize high-resolution underfoot pressure data, such as the CASIA-D \cite{Zheng2011}, SFootBD \cite{VeraRodriguez2013a}, CAD WALK \cite{Booth2018}, and UoM-Gait-69 \cite{CostillaReyes2021} datasets, usually address a limited scope, often with little or no consideration of covariates that may confound performance, and generally involve small sample sizes for subjects and walking trials, thereby under-representing real-world scenarios (see Table \ref{tab:datasets}). 
To fill this gap, UNB StepUP-P150: A footStep dataset for gait analysis and recognition using Underfoot Pressure, was developed. 
The designation P150 refers to pressure data collected from 150 participants. 
The dataset includes more than 200,000 footsteps from these 150 individuals, covering a range of walking speeds (self-paced, self-paced with a stop at the end, fast, and slow) and different types of footwear (barefoot, standard shoes, and two types of personal shoes).%, making it the most comprehensive openly available collection of its kind (Table \ref{tab:datasets}). 
The primary method for data collection involved high-resolution pressure tiles (a walkway of dimensions 1.2 meters by 3.6 meters, containing 240 by 720 pressure sensors) to capture the pressure distribution of each footstep during natural gait. 
A total of 24 minutes of walking data were collected from each participant, spread across 16 different walking conditions (4 walking speeds $\times$ 4 types of footwear) at 90 seconds per condition, resulting in approximately 1,400 steps per individual. 
Compared to the largest previously published footstep database, SFootBD, which recorded nearly 20,000 footsteps from 127 individuals \cite{VeraRodriguez2011}, with only 5 subjects contributing more than 1,000 footsteps each, the UNB StepUP-P150 dataset surpasses SFootBD by more than tenfold, making it the largest footstep dataset to date.

%%%%%%%%%%%%%%%%%%%%%%%%%%%%%%%%%%%%%%%%%%%%%%%%%%%%%%%%%%%%%%%%%%%%%%%%%%%%%%%%%%%
\begin{table}[tbh!]
\centering
\caption{Key features of public pressure-based gait databases and datasets.}
\setlength{\tabcolsep}{4pt}
\begin{threeparttable}
\footnotesize
\begin{tabular}{@{}lllllllll@{}}
\toprule
Dataset & No. IDs & \begin{tabular}[t]{@{}l@{}}No. Steps\\(Total)\end{tabular} & \begin{tabular}[t]{@{}l@{}}No. Steps\\(Per Pass)\end{tabular} & Sex & Age & Race/Ethnicity & \begin{tabular}[t]{@{}l@{}}Walking\\Speed\end{tabular} & Footwear  \\ \midrule

CASIA-D Barefoot \cite{Zheng2011} & 88\textsuperscript{a} & 2,640 & 2-4 & 66M 20F & 20-60 & Asian (Chinese) & PS, FS & BF \\

CASIA-D Shod \cite{Zheng2011} & 30\textsuperscript{b} & 540 & 2-4 & 24M 6F & 20-40 & Asian (Chinese) & PS & BF, SM  \\

SFootBD \cite{VeraRodriguez2011,VeraRodriguez2013a} & 127 & 19,980 & 2 & 83M 44F & 23.7 (18-32) & Not Specified & PS & BF, SM  \\

CAD WALK \cite{Booth2018} & 55 & 2,640 & 1 & 21M 34F & 42.3 (18-70) & White (Dutch) & PS & BF  \\
% Brian G. Booth, Noël L.W. Keijsers, Toon Huysmans, and Jan Sijbers. “The CAD WALK Healthy Controls Dataset”, June 2018, Zenodo. http://dx.doi.org/10.5281/zenodo.1265420 .

UoM-Gait-69 \cite{CostillaReyes2021} & 69 & 14,394 & 3-4 & 32M 37F & 36.4 (20-63) & Not Specified & PS, FS & SO  \\

UNB StepUP-P150 & 150 & 200,000 & 4-6 & 74M 76F & 34.2 (19-91) & White, Asian, Others & PS, FS, SS, STS & BF, SM, SC  \\

\bottomrule
\end{tabular}
\begin{tablenotes}
\footnotesize{
\textsuperscript{a} Downloadable resources include data from 96 participants, comprising roughly 2,900 footsteps. \\
\textsuperscript{b} Downloadable resources only include data from 15 participants, of whom 13 wore two different shoe types, while the remaining 2 wore only one type. \\
M: male, F: female, PS: preferred speed, FS: faster than preferred speed, SS: slower than preferred speed, STS: slow-to-stop, BF: barefoot, SO: personal shoes (one pair per participant), SM: personal shoes (multiple pairs per participant), SC: common shoes. 
}
\end{tablenotes}
\end{threeparttable}
\label{tab:datasets}

\end{table}

% Par 4: Technical Validation
The UNB StepUP-P150 dataset provides both raw trial-by-trial pressure data and preprocessed data segmented by each footstep. % , both with and without foot alignment. 
This dual format allows users to either work directly with the raw data or leverage the preprocessed data for rapid prototyping and analysis. 
To ensure its quality and consistency, the foot pressure data was subjected to thorough quality control and preprocessing, such as footstep segmentation, foot alignment, and temporal normalization. %noise filtering, and feature extraction. 
Data records are complemented by contextual metadata detailing the demographics of the participants and the conditions under which the each footstep was collected (including variables such as walking speed, footwear types, foot side, and incomplete footsteps), providing a comprehensive and fully annotated gait resource for researchers. 
In addition, video data were used for visual examination of gait events and patterns, helping to verify uncertain labels generated by automatic algorithms and providing additional context during the preprocessing and manual inspection of pressure data. 
The finalized dataset is designed for ease of use by researchers and is compatible with various analytical tools, including support of both NPZ (Python) and MAT (MATLAB) file formats.

% Par 5: Broader goals and potential reuse value
The primary aim of this work is to support the advancement of gait analysis and gait recognition, motivated by applications in biometric systems. 
To mitigate demographic biases within the dataset, the study includes participants of various ages, sexes/genders, races/ethnicities, and body sizes. %, and socioeconomic statuses. 
This ensures that the dataset accurately captures the broad variability inherent in human gait, making it both diverse and representative of the population. 
%This diversity is critical for developing generalizable models that can perform accurately across different population groups. 
In developing this gait dataset with attention to demographic bias in biometrics, the UNB StepUP-P150 dataset opens up numerous research opportunities in gait analysis and recognition, extending beyond biometric uses. 
Researchers in fields ranging from biomechanics to machine learning and deep learning can take advantage of the UNB StepUP-P150 to train and test new models, explore the relationship between underfoot pressure and other gait metrics, or investigate how various factors, such as sex, age, walking speed, or footwear, affect gait patterns. 
Furthermore, the dataset can be used as a benchmark for comparing different gait analysis and recognition techniques, thereby contributing to the standardization of methodologies in the field. 
The UNB StepUP-P150 dataset offers a level of detail and comprehensiveness that is unmatched by existing underfoot pressure-based gait datasets, making it possible to explore new avenues of research in gait analysis, footstep recognition, and related areas. 
In addition to the primary biometric focus, UNB StepUP-P150 can support research topics ranging from:

\begin{itemize}[noitemsep,topsep=0pt]
  \item statistical models of normative walking gait based on foot pressure patterns,
  \item differences in pressure-based gait patterns between demographic subgroups, categorized by factors like gender and age,
  \item differences in pressure patterns resulting from external factors such as different walking speeds and types of footwear,
  \item novel machine learning and deep learning models for gait recognition across various classification challenges,
  \item novel approaches for the segmentation, alignment, and/or registration of plantar pressure images, and
  \item development and evaluation of state-of-the-art techniques for gait analysis and gait recognition.
\end{itemize}
In summary, UNB StepUP-P150 establishes a new benchmark for plantar pressure-based gait research.
%%%%%%%%%%%%%%%%%%%%%%%%%%%%%%%%%%%%%%%%%%%%%%%%%%%%%%%%%%%%%%%%%%%%%%%%%%%%%%%%%%%
%%%%%%%%%%%%%%%%%%%%%%%%%%%%%%%%%%%%%%%%%%%%%%%%%%%%%%%%%%%%%%%%%%%%%%%%%%%%%%%%%%%
%%%%%%%%%%%%%%%%%%%%%%%%%%%%%%%%%%%%%%%%%%%%%%%%%%%%%%%%%%%%%%%%%%%%%%%%%%%%%%%%%%%
\section*{Methods}

%%% The Methods should include detailed text describing any steps or procedures used in producing the data, including full descriptions of the experimental design, data acquisition assays, and any computational processing (e.g. normalization, image feature extraction). See the detailed section in our submission guidelines for advice on writing a transparent and reproducible methods section. Related methods should be grouped under corresponding subheadings where possible, and methods should be described in enough detail to allow other researchers to interpret and repeat, if required, the full study. Specific data outputs should be explicitly referenced via data citation (see Data Records and Citing Data, below). %%% Authors should cite previous descriptions of the methods under use, but ideally the method descriptions should be complete enough for others to understand and reproduce the methods and processing steps without referring to associated publications. There is no limit to the length of the Methods section. Subheadings should not be numbered.

%%%%%%%%%%%%%%%%%%%%%%%%%%%%%%%%%%%%%%%%%%%%%%%%%%%%%%%%%%%%%%%%%%%%%%%%%%%%%%%%%%%
%%%%%%%%%%%%%%%%%%%%%%%%%%%%%%%%%%%%%%%%%%%%%%%%%%%%%%%%%%%%%%%%%%%%%%%%%%%%%%%%%%%
\subsection*{Participants}

A total of 180 individuals participated in the study, which took place over a period of 18 months at the University of New Brunswick (UNB) in Fredericton, New Brunswick, Canada. 
The participants consisted of students, staff, and faculty from the university, and community members who were recruited via word-of-mouth and promotional materials circulated to local businesses and clubs. 
The general inclusion criteria were: (1) at least 18 years of age; (2) able to walk comfortably without assistive devices for 90 seconds at a time with different walking speeds and footwear conditions (24 minutes total); and (3) able and comfortable to balance unassisted, on both feet and on one foot, for 30 seconds at a time with different footwear conditions (6 minutes total). % ; and (4) able to walk and stand without any physical or medical condition for which the protocol would be contraindicated. 
Before collecting data, all participants provided their written informed consent for their participation and for the inclusion of their data in the public dataset release. 
The study protocol was reviewed and approved by the University of New Brunswick's Research Ethics Board (REB 2022-132), in accordance with the Declaration of Helsinki.

The following factors led to participant exclusion from the final dataset to maintain its quality and integrity: (1) experimental deviations, e.g. unintended behaviour during one or more trials ($N = 2$); (2) missing data, e.g. incomplete protocols or corrupted files ($N = 4$); (3) hardware malfunction, e.g. disconnected sensors significantly affecting recordings ($N = 15$), and (4) no accompanying video data to confirm annotations, as participants did not consent to video collection ($N = 9$). % Note: Some experiments included in our 150 do have minor hardware issues (e.g., minimal number of passes and trials affected), but here we exclude those where data is largely impacted (multiple trials affected, and/or several tiles disconnected during a trial). % Note: we did not specifically exclude participants due to injury, disability, or other mobility conditions, however we asked them to disclose conditions that may impact their gait.
Consequently, 30 participants were removed from the final dataset, resulting in a total of 150 subjects being included. 
None of the participants reported having a neurological condition (e.g., stroke or Parkinson's), a concussion within six months of recording, or pregnancy.
Moreover, although not made publicly available as part of the UNB StepUP-P150 dataset, portions of the data from the 30 excluded experiments will be used to support future footstep biometric competitions as independent test samples.

\begin{figure}[!tb]
    \centering
    \includegraphics[width = \textwidth]{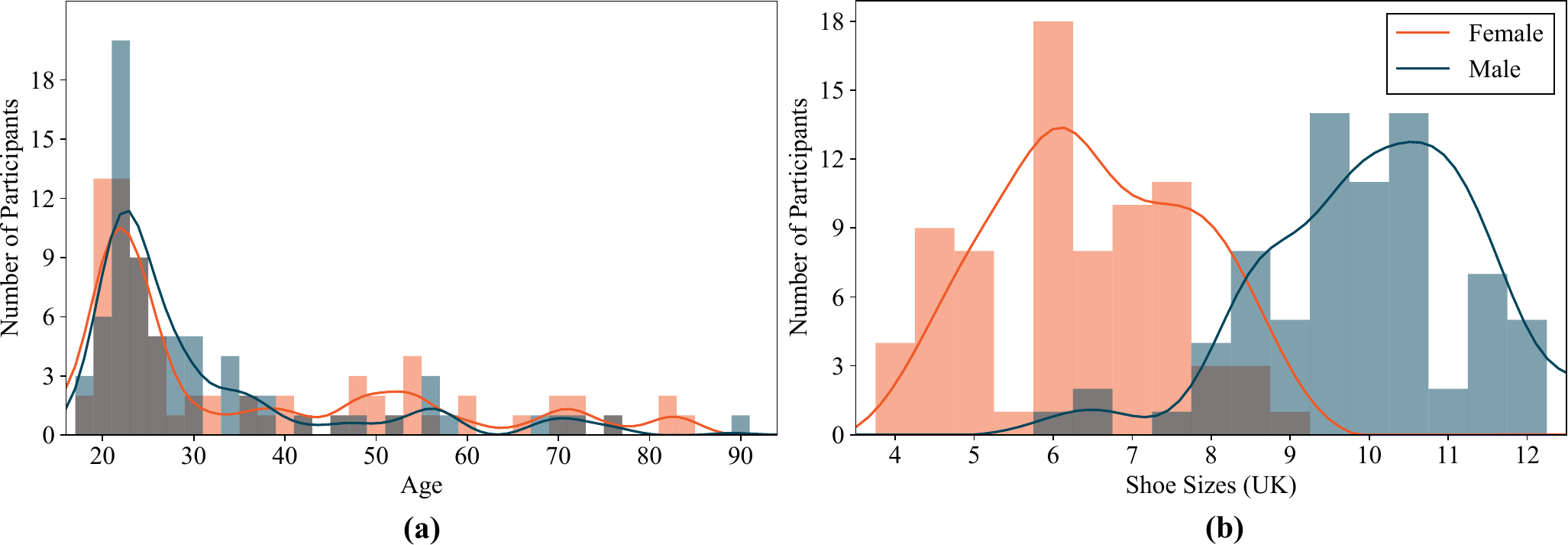}
    \caption{Distributions of (a) participants' ages by sex and (b) participants' chosen standard shoe sizes by sex, in UK sizes. Note: Female distributions are shown in orange, and male distributions are shown in dark blue. A Wilcoxon rank-sum test did not indicate any significant difference in the distribution of ages for the female and male subgroups ($p = 0.55$). The shoe sizes for the male subgroup were significantly larger than the female subgroup ($p < 0.0001$ using a two sample $t$-test).}
    \label{fig:histogram}
\vspace{-2mm}
\end{figure}

\begin{figure}[!tb]
    \centering\includegraphics[width=\textwidth]{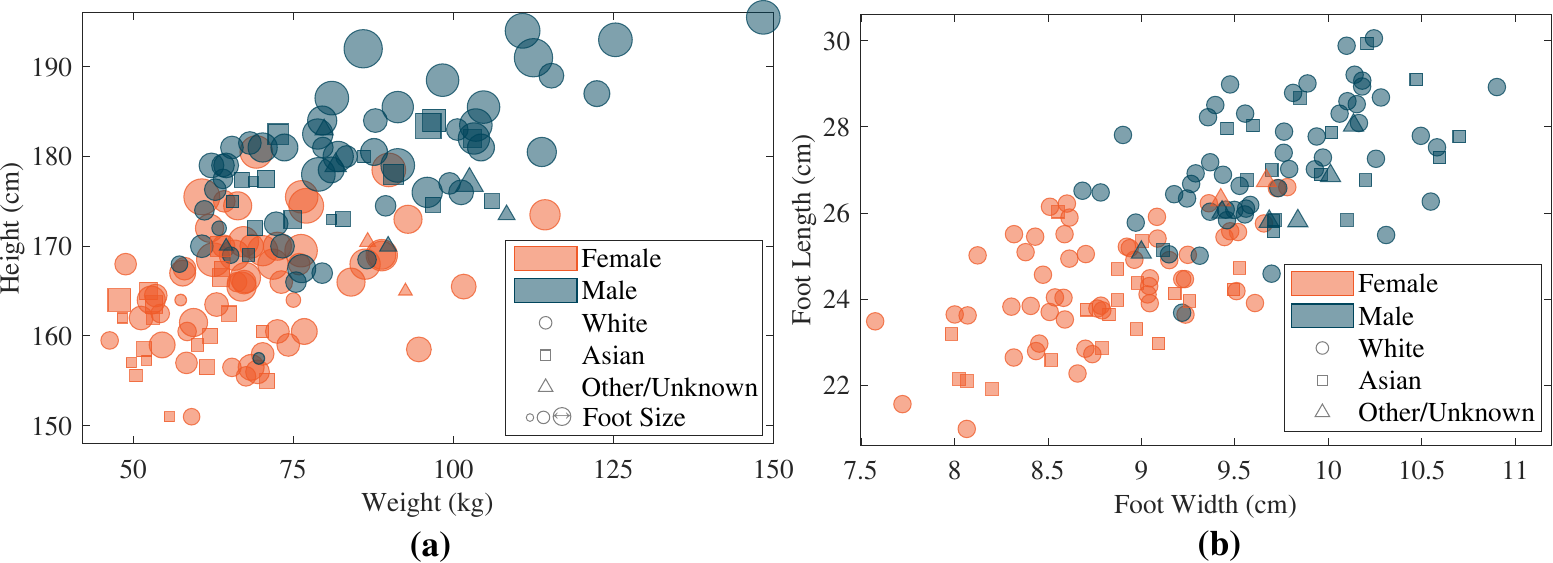}
    \caption{Distributions of participants' physical characteristics by sex and ethnicity/race: (a) height (cm), weight (kg), and foot size (marker size is proportional to measured foot length in cm), and (b) measured foot length (cm) and width (cm). Note: Orange markers are used for female participants and dark blue for male participants. Some jitter was added for (b) to improve visibility.}
    \label{fig:anthropometric}
\vspace{-4mm}
\end{figure}

Table \ref{tab:demographics} provides a summary of the demographic and anthropometric data for the 150 participants in the study. 
The dataset includes 74 males and 76 females with an average age of 34 years, spanning 19 to 91 years. 
The dataset maintains an approximately balanced sex/gender distribution and includes both younger and older adults, with the age distribution for participants by sex illustrated in Fig.~\ref{fig:histogram}(a). 
Most of the participants identify as White ($N=106$, $71\%$), with additional identifications as Asian ($N=36$, $24\%$; 15 Middle Eastern, 10 South Asian, 11 East/Southeast Asian), Other/Multiple ($N=6$, $4\%$; 1 Black, 5 multi-ethnic), and Unknown/Not Specified ($N=2$, $1\%$). 
This reflects an over-representation of visible minority groups relative to the official population distribution in the local community; the proportion of total visible minorities in the database is 28\%, as opposed to the 14\% reported by Statistics Canada \cite{census2023}. 
The dataset also includes participants with a variety of body types and dimensions, including height ranging from 151 to 196 cm, weight between 46 and 148 kg, BMI spanning 17 to 39 kg/m\textsuperscript{2}, foot length measuring 20 to 30 cm, foot width between 7 and 11 cm, and UK shoe sizes from 4 to 12.5 (Fig. \ref{fig:histogram}(b)). 
Figure \ref{fig:anthropometric} shows the distribution of participants' body sizes and foot measurements, categorized by both sex and ethnicity/race.

%%%%%%%%%%%%%%%%%%%%%%%%%%%%%%%%%%%%%%%%%%%%%%%%%%%%%%%%%%%%%%%%%%%%%%%%%%%%%%%%%%%
\begin{table}[!tb]
\centering
\footnotesize
\caption{\label{tab:demographics}Demographics and physical characteristics of the participants included in the UNB StepUP-P150 dataset.}
\setlength{\tabcolsep}{35pt}

\begin{tabular}{@{}ll@{}}
\toprule
Factor                       & \multicolumn{1}{c}{Number or $\mu \pm \sigma$ (range)} \\ \midrule
\textit{Age}                    & $34.2 \pm 17.3$ (19 – 91)   \\
\textit{Height (cm)}            & $171.5 \pm 9.6$ (151 - 196) \\
\textit{Weight (kg)}            & $76.0 \pm 18.4$ (46 - 148)  \\
\textit{BMI (kg/m\textsuperscript{2})} & $25.6 \pm 4.7$ (17 - 39)                                             \\
\textit{Foot Length (cm)}       & $25.7 \pm 2.0$ (20 - 30)    \\
\textit{Foot Width (cm)}        & $9.3 \pm 0.7$ (7 - 11)      \\
\textit{Shoe Size (UK)}         & $8.0 \pm 2.2$ (4 - 12.5)      \\
\textit{Preferred Walking Speed (m/s) }& $1.12 \pm 0.15$             \\
\textit{Race/Ethnicity}         &                             \\
\quad White                     & 106                         \\
\quad Asian                     & 36                          \\  
\quad Other/Multiple            & 6                           \\
\quad Unknown/Not Specified     & 2                           \\
\textit{Sex}                    &                             \\
\quad Male                      & 74                          \\
\quad Female                    & 76                          \\
\textit{Gender}                 &                             \\
\quad Man                       & 72                          \\
\quad Woman                     & 75                          \\
\quad Non-Binary                & 1                           \\
\quad Not Specified             & 2                           \\
\textit{Dominant Leg}           &                             \\
\quad Right                     & 140                         \\
\quad Left                      & 10                          \\ 
\bottomrule
\end{tabular}
\end{table}

%%%%%%%%%%%%%%%%%%%%%%%%%%%%%%%%%%%%%%%%%%%%%%%%%%%%%%%%%%%%%%%%%%%%%%%%%%%%%%%%%%%

%%%%%%%%%%%%%%%%%%%%%%%%%%%%%%%%%%%%%%%%%%%%%%%%%%%%%%%%%%%%%%%%%%%%%%%%%%%%%%%%%%%

%%%%%%%%%%%%%%%%%%%%%%%%%%%%%%%%%%%%%%%%%%%%%%%%%%%%%%%%%%%%%%%%%%%%%%%%%%%%%%%%%%%
\subsection*{Instrumentation}

\begin{figure}[!tb]
\centering
    \includegraphics[width=0.8\textwidth]{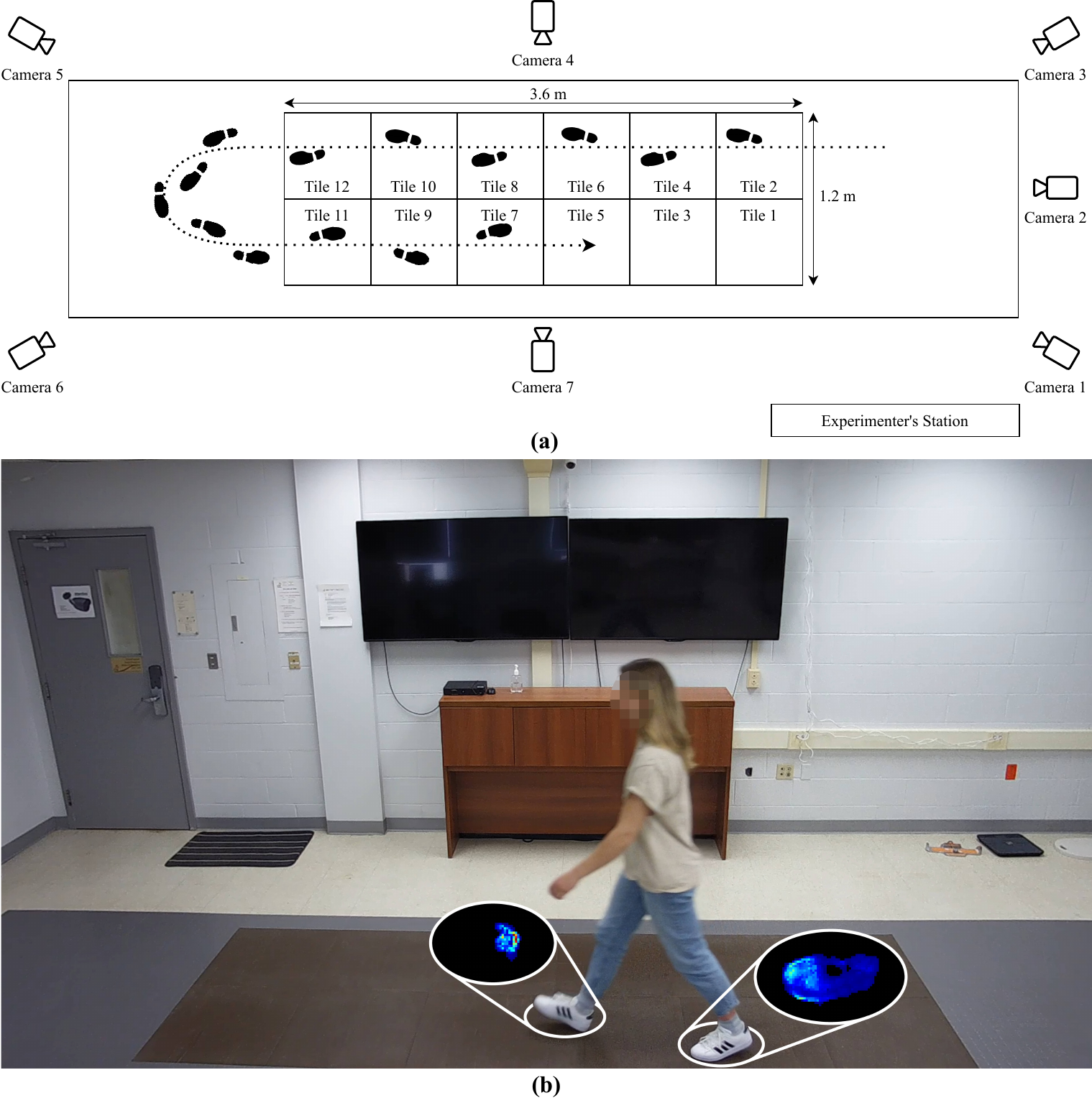}
    \caption{Overview of the instrumentation configuration. (a) A diagram of the laboratory setup; participants walked back and forth across a grid of twelve sensing tiles encircled by a non-instrumented platform to allow for turning. Seven RGB video cameras were used to capture the participants from different viewing angles. (b) A video frame from Camera 7 during a walking trial with corresponding pressure measurements.}
    \label{fig:sensor_setup}
\vspace{-4mm}
\end{figure}

% tiles
Pressure data from footsteps were collected using a specialized runway (Fig. \ref{fig:sensor_setup}), which featured a $2\times6$ grid of commercial pressure sensing tiles developed by Stepscan Technologies Inc. 
The runway has a total active recording area of 1.2 m $\times$ 3.6 m, allowing for the collection of multiple consecutive steps (typically 4-6 steps). 
Each modular 60 cm $\times$ 60 cm tile is made up of a $120 \times 120$ sensor grid, corresponding to a spatial resolution of 4 sensors/cm$^2$. 
Collectively, the $2 \times 6$ tile grid comprises 172,800 piezoresistive sensors (i.e., $240 \times 720$), each with a threshold sensitivity of 10 kPa, sensing range of 1,510kPa (10 kPa resolution), and 100 Hz sampling rate.
To increase system durability, particularly in areas experiencing high foot traffic and the use of outdoor footwear, as anticipated in the practical application of this study, the manufacturer tailored the tiles with a 6 mm protective layer made of Ramflex double-layered vulcanized rubber flooring, instead of the usual 2 mm thickness commonly applied in clinical usage of this technology for healthcare applications. 
These tiles were installed in a laboratory reserved for data collection and stayed there throughout the entire study. 
A non-instrumented wooden platform was also constructed around the tile grid, extending 1.4 meters at each end of the runway. 
The platform was flush with the tiles to allow natural walking behaviour, eliminating the need to step up or down and providing room for the participants to turn during the experiment. 
It should be noted that the pressure-sensitive tiles are pre-calibrated by the manufacturer (Stepscan Technologies Inc.), and did not require on-site calibration. 
Each tile comes with factory-set parameters that align the signal characteristics of each physical pressure sensor with a common logical signal domain. 
The companion data collection application automatically implements these corrections, ensuring that they are reflected in all data within this dataset. 
Therefore, no further calibration efforts were undertaken by the research team.

% cameras, % scanner
In addition to pressure-sensing tiles, seven RGB video cameras (QCN8068BA, Q-See, USA) with a resolution of 1080p ($1920\times1080$) and a frame rate of 20 fps were strategically installed around the room. % Note: we lowered the resolution from 1440p to 1080p (1920 x 1080) for space & speed
These cameras were positioned to record participants' gait from various angles: approximately 0\textdegree, 45\textdegree, 90\textdegree, 135\textdegree, 225\textdegree, 270\textdegree, and 315\textdegree~with respect to the major axis of the tile grid (Fig. \ref{fig:sensor_setup}(a)). 
Participants in the UNB StepUP-P150 dataset consented to have their sessions videotaped, producing seven separate video recordings for each pressure data capture. 
In addition, a flatbed scanner (OpticSlim 1180, Plustek, USA) with an optical resolution of 300 dpi, 48-bit color input, and a scanning area of 29.7 cm $\times$ 43.18 cm was used to acquire digital scans of the participant's two pairs of personal footwear that they brought to their session. % Note: we have it set to optical resolution of 300p (they only recommend 100p for color, for speed & space) (previously 1200p)
The pressure and video recordings were captured simultaneously on two networked desktop computers, connected via the Secure Shell (SSH) protocol, and time-synchronized using the Network Time Protocol (NTP). 
Each recording was time-stamped to enable precise offline synchronization of the data from the two sensor types.
Although video recordings and digital shoe scans are not included in the StepUP-P150 release, these were used internally by the research team to manually confirm metadata and pressure data integrity (see Technical Validation, below).

%%%%%%%%%%%%%%%%%%%%%%%%%%%%%%%%%%%%%%%%%%%%%%%%%%%%%%%%%%%%%%%%%%%%%%%%%%%%%%%%%%%

%%%%%%%%%%%%%%%%%%%%%%%%%%%%%%%%%%%%%%%%%%%%%%%%%%%%%%%%%%%%%%%%%%%%%%%%%%%%%%%%%%%
%%%%%%%%%%%%%%%%%%%%%%%%%%%%%%%%%%%%%%%%%%%%%%%%%%%%%%%%%%%%%%%%%%%%%%%%%%%%%%%%%%%
\subsection*{Experimental Protocol}

\begin{figure}[!tb]
\begin{center}
\includegraphics[width=\linewidth]{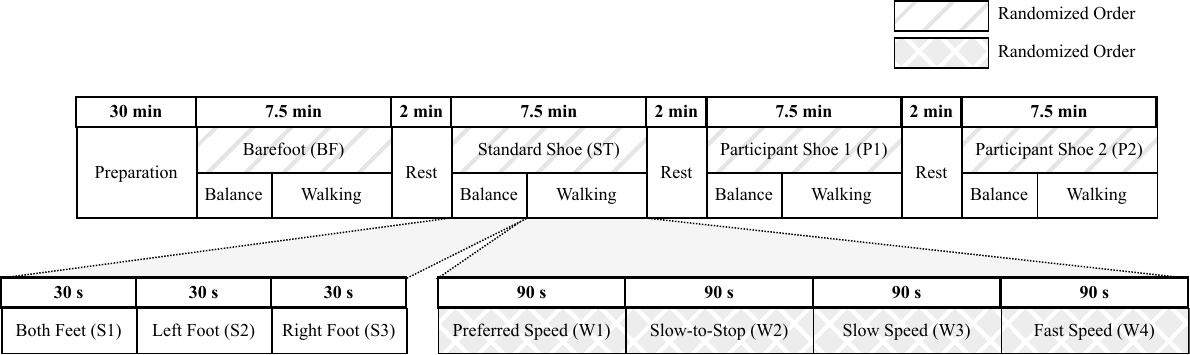}
\end{center}
% \vspace{-5mm}
\caption{Overview of the experimental protocol. After a 30 minute preparation period for onboarding and familiarizing the participant with the study, three 30-second standing trials (S1, S2, and S3) and four 90-second walking trials (W1, W2, W3, and W4) were recorded for each of the four footwear conditions (BF, ST, P1, and P2). The participants were allowed to take breaks throughout the study as needed, with at least two minutes taken to sit down and change shoes between footwear conditions. The order of the footwear conditions and walking speeds were randomized for each participant.}
\label{fig:protocol_tasks}
\vspace{-5mm}
\end{figure}

Figure~\ref{fig:protocol_tasks} provides a summary of the 90-minute experimental protocol. 
The initial preparation session took around 30 minutes, followed by about an hour of balance and walking sessions. 
Before starting the walking trials, three separate 30-second balance tests were performed for each type of footwear. 
The walking experiment then involved sixteen 90-second walking trials, featuring four different footwear conditions, each executed at four distinct walking speeds. 
As a result, there were seven recorded trials for each type of footwear, resulting in 28 trials overall and a total recording duration of 30 minutes. 
Participants were allowed to take breaks and rest between trials as needed, and whenever there was a need to change shoes, a minimum of 2 minutes was allocated for rest. 
To minimize the potential impact of fatigue or ordering effects on the recorded gait patterns, the order of footwear and walking speed conditions was randomized for each participant (see Fig. \ref{fig:protocol_tasks} for an example sequence). 
The subsequent sections provide a more detailed explanation of different tasks and conditions.

%%%%%%%%%%%%%%%%%%%%%%%%%%%%%%%%%%%%%%%%%%%%%%%%%%%%%%%%%%%%%%%%%%%%%%%%%%%%%%%%%%%

%%%%%%%%%%%%%%%%%%%%%%%%%%%%%%%%%%%%%%%%%%%%%%%%%%%%%%%%%%%%%%%%%%%%%%%%%%%%%%%%%%%
\subsubsection*{Preparation of the Participant}

Before the scheduled collection session, the participants were informed that the session would consist of a series of walking and balance tasks, and they were asked to bring along two pairs of their own shoes for the experiment. 
Upon their arrival, participants received an overview of the experimental procedure. 
They then filled out a form to provide essential demographic details, including sex, gender, birth year, race or ethnicity (options included Aboriginal, Black, East/Southeast Asian, Latino, Middle Eastern, South Asian, White, or Other), along with pertinent physical information that could influence their gait patterns, such as any recent injuries and leg dominance, determined by answering ``Which foot would you normally use to kick a stationary ball straight in front of you?''\cite{vanMelick2017}. 
To obtain precise anthropometric measurements, the experimenter assisted each participant in recording their height, body weight, and foot size. 
Body weight was measured using a Withings body smart scale, height with a Seca 213 portable stadiometer, and foot length and width with an adult Brannock device. 
In addition, digital shoe sole scans were obtained from the two pairs of participant-provided footwear, with the sizes, brands, and descriptions of each shoe documented. 
The participants were then fitted with a pair of standard shoes supplied by the research team, specifically Adidas Grand Court 2.0 unisex sneakers, available in sizes from US Men's size 4 (Women's size 5) to US Men's size 13 (Women's size 14), inclusive of half sizes. 
This is equivalent to UK sizes 3.5 to 12.5.
Participants were instructed to select the size that offered the most comfort and to take a few steps in the shoes to verify fit. 
Their preferred sizes were thereafter documented.

%%%%%%%%%%%%%%%%%%%%%%%%%%%%%%%%%%%%%%%%%%%%%%%%%%%%%%%%%%%%%%%%%%%%%%%%%%%%%%%%%%%
\subsubsection*{Balance Tasks}

Before each balance and walking trial, participants received a brief reminder about the desired task. 
The recording software emitted an audible ``beep'' to signify the beginning of the trial and participants were instructed to continue their task until the experimenter signaled the end of the trial. 
For the three balance tasks, the participants stood for 30 seconds at a time on \textbf{both feet (S1)}, on their \textbf{left foot (S2)}, and on their \textbf{right foot (S3)}. 
They were asked to select one location on the tiles and to remain stationary for the duration of the recording and were not permitted to use any external supports during these trials (e.g., holding onto a chair). 
If they lost balance during recording, they were asked to re-adjust as necessary and attempt to continue the task until the 30 seconds were complete.

%%%%%%%%%%%%%%%%%%%%%%%%%%%%%%%%%%%%%%%%%%%%%%%%%%%%%%%%%%%%%%%%%%%%%%%%%%%%%%%%%%%
\subsubsection*{Walking Tasks: Walking Speed}

For the walking trials, the participants were instructed to walk back and forth across the tiles naturally for 90 seconds along the longer 3.6 m grid direction, making use of the non-instrumented portion of the platform to turn around (Fig. \ref{fig:sensor_setup}). 
The four walking speed conditions were:
\begin{itemize}
\item \textbf{Preferred Speed (W1)}: Participants were instructed to walk at a natural, self-selected (moderate) pace that was comfortable to them. 
\item \textbf{Slow-to-Stop (W2)}: Participants were instructed to walk at their preferred speed, slowing to an abrupt, two-foot stop at the end of each pass on the tile grid, maintaining this stop for approximately 1 second. This condition was designed to mimic the walking behaviour of approaching a controlled access point or security turnstile (e.g., metro, secured office building), focusing on capturing changes in walking speed. 
\item \textbf{Slow (W3)}: Participants were instructed to walk at a pace slower than their naturally chosen speed (self-selected slower walking).
\item \textbf{Fast (W4)}: Participants were instructed to walk at a pace faster than their naturally chosen speed (self-selected faster walking). 
\end{itemize}

\begin{figure}[!h]
\centering
\includegraphics[width=0.5\textwidth]{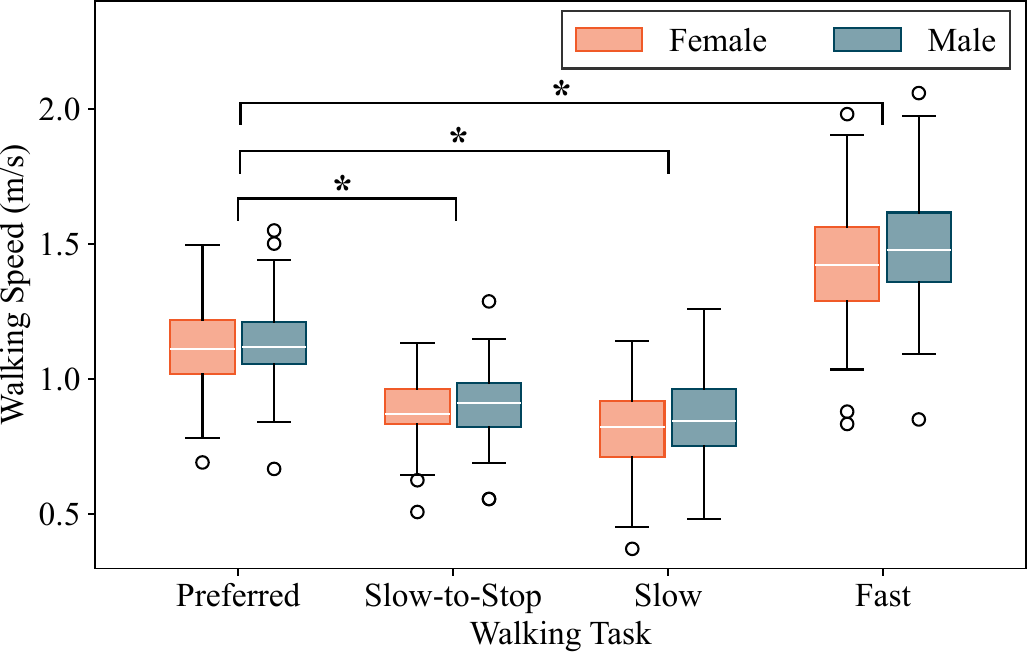}
\caption{Average walking speeds, categorized by sex, as computed from pressure measurements during each walking task of the experimental procedure, and averaged over the four different footwear conditions. Note: The slow-to-stop, slow, and fast walking speeds were all found to be significantly different than the participants' preferred walking speeds ($p < 0.05$ using paired $t$-tests). There were no statistically significant differences between walking speeds for female and male subgroups ($p > 0.05$ for all walking tasks using two-sample $t$-tests)} % all comparisons of four walking speeds to each other were significantly different
    \label{fig:speed_boxplot}
\vspace{-3mm}
\end{figure}

The average self-selected preferred (moderate) pace was 1.12 m/s, the average fast speed was 1.45 m/s (which is $26\%$ faster) and the average slow speed was 0.83 m/s (which is $29\%$ slower) (see Fig.~\ref{fig:speed_boxplot}). 
It should be noted that during the trials, the participants had the freedom to choose any walking path and to make the 180-degree turns between passes in any way they preferred. 
However, the experimenters advised the participants to change their turning directions and/or pause on the non-instrumented landing when necessary to reduce the possibility of dizziness.

%%%%%%%%%%%%%%%%%%%%%%%%%%%%%%%%%%%%%%%%%%%%%%%%%%%%%%%%%%%%%%%%%%%%%%%%%%%%%%%%%%%

%%%%%%%%%%%%%%%%%%%%%%%%%%%%%%%%%%%%%%%%%%%%%%%%%%%%%%%%%%%%%%%%%%%%%%%%%%%%%%%%%%%
\subsubsection*{Walking Tasks: Footwear}

\begin{figure}[!tb]
\begin{center}
\includegraphics[width=0.85\linewidth]{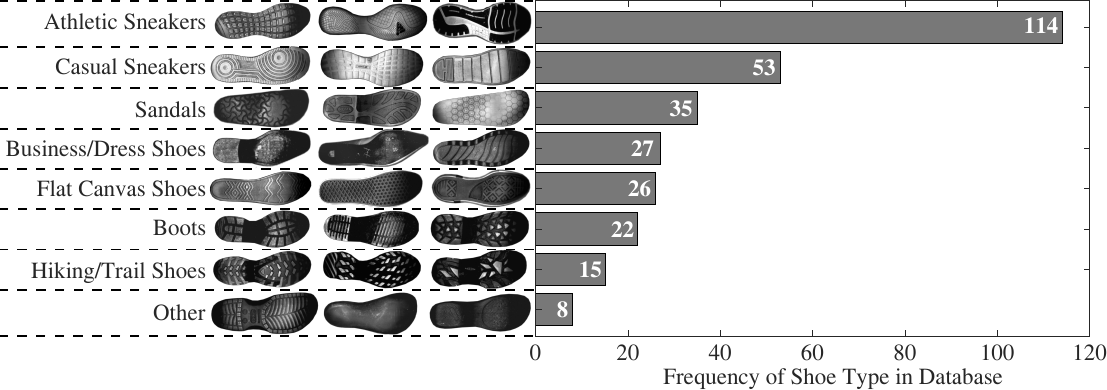}
\end{center}
\vspace{-5mm}
\caption{Distribution of participants' personal footwear types, complemented by digital scan examples for each category.}
\label{fig:shoe_histogram}
% \vspace{-5mm}
\end{figure}

The trials were completed with four different footwear conditions:
\begin{itemize}
\item \textbf{Barefoot (BF)}: without shoes. Participants had the option to conduct these trials either with socks ($N = 114$) or completely barefoot ($N = 36$). This decision is recorded in the \textit{BFType} metadata field. 
\item \textbf{Standard Shoe (ST)}: a common pair of flat-soled casual sneakers (Adidas Grand Court 2.0) provided by the research team. 
\item \textbf{Personal Shoes (P1 and P2)}: two pairs of personal shoes frequently worn by the participants. The personal shoes included a wide range of different shoe types, which have been broadly categorized as athletic sneakers (e.g., road running shoes, indoor trainers), casual sneakers (e.g., skate shoes, court-inspired sneakers, high-top leather sneakers), sandals (e.g., slides, flip-flops, ankle-strap sandals), flat canvas shoes (e.g., lace-up or slip-on shoes with canvas upper), boots (e.g., winter boots, steel-toe work boots, Chelsea boots), business/dress shoes (e.g., high heels, ballet flats, oxfords), hiking/trail shoes (e.g., trail running shoes, outdoor walking shoes), and other (e.g., foam clogs, minimal shoes). Figure \ref{fig:shoe_histogram} presents the breakdown of shoes brought in by the participants, grouped into these eight categories. The most commonly brought shoes were athletic sneakers ($N = 114$), followed by casual sneakers ($N = 53$) and sandals/flip flops ($N = 35$). 
A smaller number of shoes with unique pressure patterns, such as high heels ($N = 3$) and steel or composite toe work boots ($N = 5$), were also included.
\end{itemize}

%%%%%%%%%%%%%%%%%%%%%%%%%%%%%%%%%%%%%%%%%%%%%%%%%%%%%%%%%%%%%%%%%%%%%%%%%%%%%%%%%%%
%%%%%%%%%%%%%%%%%%%%%%%%%%%%%%%%%%%%%%%%%%%%%%%%%%%%%%%%%%%%%%%%%%%%%%%%%%%%%%%%%%%

\begin{figure}[!tb]
\begin{center}
\includegraphics[width=0.5\linewidth]{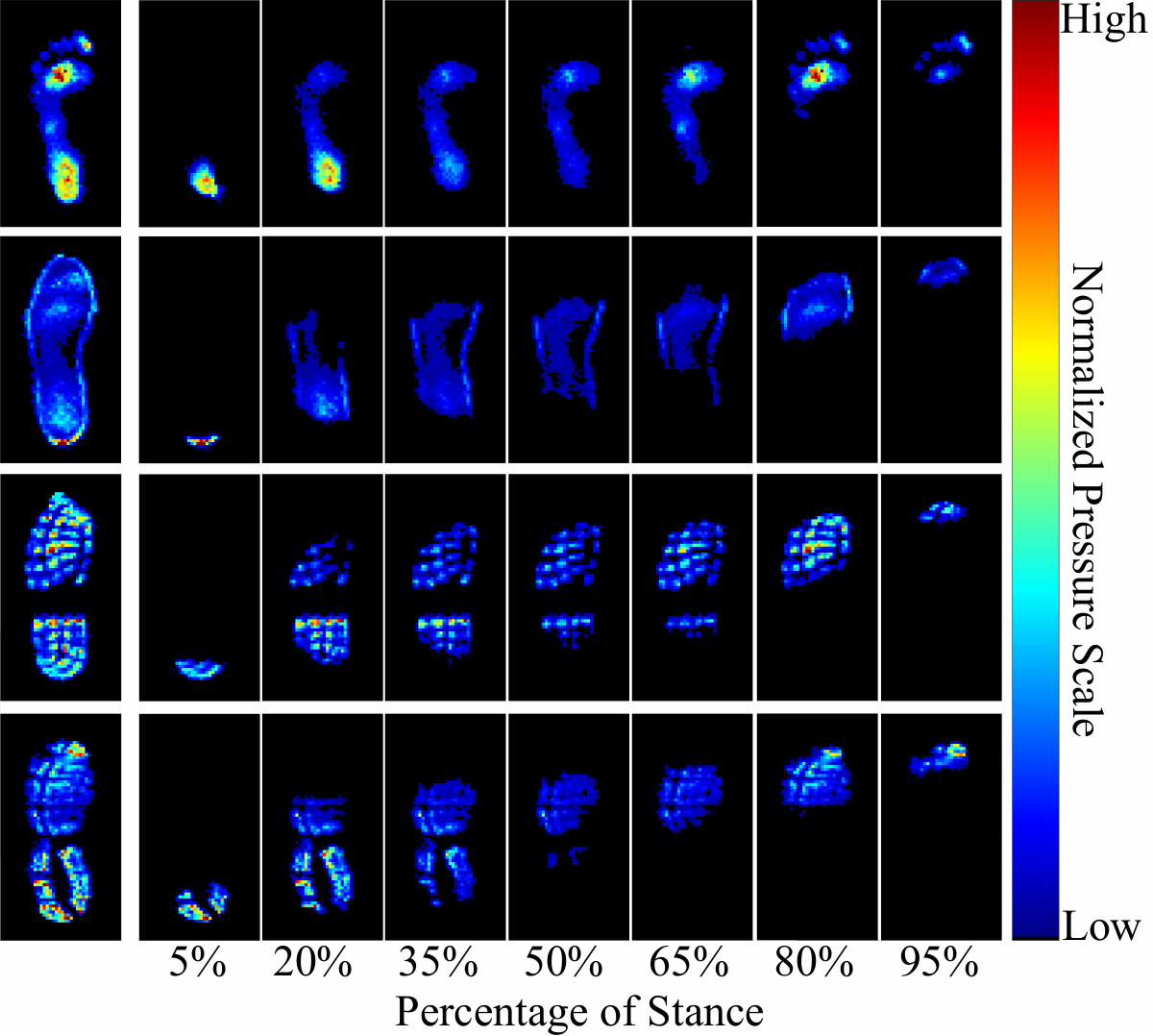}
\end{center}
\vspace{-5mm}
\caption{Example pressure image time series from the same participant, (1) without footwear (top row), (2) wearing standard shoes (second row), (3) wearing a pair of the participant's personal work boots (third row), and (4) wearing a pair of the participant's personal running shoes (last row), plotted at multiple phases throughout the stance.}
\label{fig:footstep_progression}
\vspace{-4mm}
\end{figure}

\begin{figure}
    \centering
     \includegraphics[width=\textwidth]{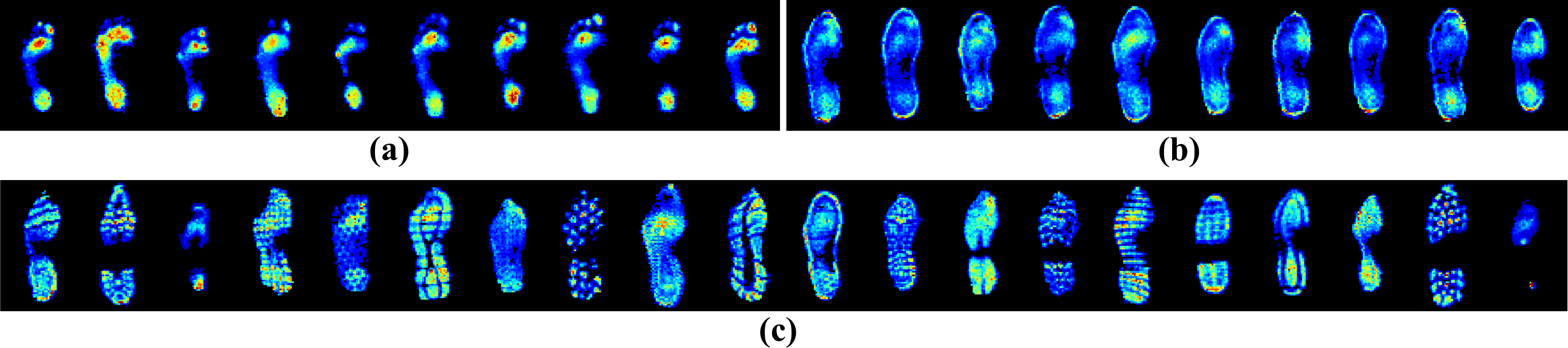}
    \caption{Example peak pressure images for a selection of footsteps; (a) without footwear (from the BF trials), (b) wearing standard shoes (from the ST trials), and (c) wearing various types of personal footwear (from the P1 and P2 trials), including steel-toe work boots (second from right) and stiletto high heels (right).}
    \label{fig:P100s}
\vspace{-4mm}
\end{figure}

Figure \ref{fig:footstep_progression} shows the time series of pressure images for an individual taking a single footstep, highlighting changes across the stance phase in four different footwear conditions, whereas Figure \ref{fig:P100s} illustrates variations in peak pressure profiles across different participants and shoe types.

%%%%%%%%%%%%%%%%%%%%%%%%%%%%%%%%%%%%%%%%%%%%%%%%%%%%%%%%%%%%%%%%%%%%%%%%%%%%%%%%%%%

%%%%%%%%%%%%%%%%%%%%%%%%%%%%%%%%%%%%%%%%%%%%%%%%%%%%%%%%%%%%%%%%%%%%%%%%%%%%%%%%%%%

%%%%%%%%%%%%%%%%%%%%%%%%%%%%%%%%%%%%%%%%%%%%%%%%%%%%%%%%%%%%%%%%%%%%%%%%%%%%%%%%%%%
\subsection*{Data Processing}

A total of 28 raw pressure data recordings were captured for each participant, twelve 30-second balance recordings (three recordings for each footwear condition) and sixteen 90-second walking recordings (four recordings for each footwear condition). 
Each raw data recording is stored as a 3D tensor with dimensions of nominally 3000 frames $\times$ 720 pixels $\times$ 240 pixels and 9000 frames $\times$ 720 pixels $\times$ 240 pixels, for the 30-second and 90-second trials respectively. 
It should be noted that the duration of trials may differ slightly by a few frames (typically less than 100 frames, equivalent to about one second) as a result of minor delays when beginning or ending the recordings. 
The recordings consist of unprocessed, trial-by-trial data capturing continuous standing or walking on the pressure-sensitive tiles. 
These raw pressure recordings were processed to offer a user-friendly labeled data format suitable for both gait analysis and gait recognition, ensuring quality and consistency.   
The subsequent sections provide a more detailed explanation of the preprocessing steps for the balance and walking trials.

%%%%%%%%%%%%%%%%%%%%%%%%%%%%%%%%%%%%%%%%%%%%%%%%%%%%%%%%%%%%%%%%%%%%%%%%%%%%%%%%%%%
\subsubsection*{Balance Trial Preprocessing}

During the balance trials, participants were allowed to stand facing any direction and choose any tile(s) to stand on. 
Therefore, for ease of analysis, the balance trial recordings were (1) cropped to focus on the region of interest (ROI) and (2) adjusted to share a common orientation. 
In particular, the areas of activity on the tiles within the 30-second recordings (reflecting pressures from one or both feet) were detected using a threshold of 10 kPa. 
These areas were then extracted from the sensor grid and zero padded to fit a 2D tensor of dimensions $180 \times 180$ pixels. 
To ensure that all recordings are approximately aligned, with the big toe of the right foot directed toward the top left corner and the big toe of the left foot toward the top right corner, the cropped recordings were rotated in 90-degree increments as needed, based on visual inspection.
Temporally, the recordings were cropped by a few frames at the beginning of the trial to a total duration of 3000 frames.
No additional normalization in terms of space, time, or amplitude was performed, resulting in a 3D tensor of 3000 frames $\times$ 180 pixels $\times$ 180 pixels for each trial.

%%%%%%%%%%%%%%%%%%%%%%%%%%%%%%%%%%%%%%%%%%%%%%%%%%%%%%%%%%%%%%%%%%%%%%%%%%%%%%%%%%%
\subsubsection*{Footstep Detection and Extraction}

For the walking trials, individual steps were identified both spatially and temporally within the raw pressure data to enable analysis on a step-by-step basis. 
%Object detection involves determining an object's location and outline, capturing both its center and boundary.
%It also includes classifying the object, which, in this context, determines whether or not it is a footstep. 
Footstep detection and tracking in this study were achieved through simple yet efficient techniques, which present possibilities for real-time online implementation during system operation. 
Starting from the first frame, the regions of activity on the tiles at each time point of the recording, potentially indicating contact from one or both feet, were identified using a simple connected pixels object detection technique. 
Specifically, the $720 \times 240$ frames of the recording were first converted to binary images using a threshold of 10 kPa. 
The resulting binary frames were then processed with morphological operations: dilation using a circular structuring element with a radius of 4, followed by erosion with a circular element of radius 2. 
Subsequently, pixels that were 2-connected (i.e., separated by no more than two orthogonal hops) were clustered to form objects. 
In this research, objects whose centroids were within a specified distance (e.g., 20 pixels) were combined since they frequently represented the heel and forefoot of a shoe or foot with a high arch. 
Adjustments to this distance were made as needed to accommodate certain special sole types, including stiletto high heels. 

After identifying the bounding boxes for objects within a frame, SORT (Simple Online and Realtime Tracking) \cite{Bewley2016} was used to track these objects across subsequent frames. 
SORT uses a Kalman filter combined with a linear motion model to predict object locations based on prior positions, linking bounding boxes over time by assessing the overlap between predicted and observed positions.
This method of footstep extraction produced 3D bounding boxes with dimensions (time, height, width), or ($t, y, x$), for each step, where $x$ and $y$ depend on the size and orientation of the participant's foot, and $t$ changes according to walking speed. 
These bounding boxes are supplied as footstep metadata. 
Since each pass over the tiles records several consecutive steps, they enable the computation of various spatiotemporal gait parameters including gait speed, step length, step width, stance time, swing time, step angle, and walking path, among others.

%%%%%%%%%%%%%%%%%%%%%%%%%%%%%%%%%%%%%%%%%%%%%%%%%%%%%%%%%%%%%%%%%%%%%%%%%%%%%%%%%%%
\subsubsection*{Footstep Normalization}

To enable more advanced analytic methods, including machine learning and deep learning, it is recommended to store footstep data in tensors, which are essentially multidimensional arrays. 
Since the shape of each footstep varies due to factors such as foot size, foot angle, and walking speed, padding techniques can be employed to standardize spatial and/or temporal dimensions across different steps. 
Applying zero padding helps preserve information related to direct factors, such as foot size and walking speed, within the dataset, which may provide significant information for classification \cite{JlassiDixon2024}. 
While these methods allow the 3D footstep tensors to be organized into arrays (i.e., higher-dimensional tensors), they do not support meaningful comparisons of plantar pressure across consistent foot regions, gait events, or pressure intensity scales.
In the literature on gait analysis, normalization techniques have been applied in three distinct domains: spatial normalization (e.g., rotating for the foot progression angle and scaling for foot size), temporal normalization (e.g., interpolating to either 100 or 101 frames to represent the full range from $0\%$ to $100\%$ of the gait cycle), and amplitude normalization (e.g., scaling according to body mass or total foot pressure). 
Although these methods allow for comparisons of footsteps in both spatial and temporal domains, crucial information about subject characteristics might be lost. 
Consequently, it has been shown that merging various preprocessing pipelines may enhance the efficacy of footstep recognition systems \cite{CostillaReyes2019}.

Given the absence of a universally accepted preprocessing pipeline suitable for all deep learning models and their specific classification or recognition tasks \cite{JlassiDixon2024}, the UNB StepUP-P150 dataset provides two different versions of the extracted footsteps as examples, each demonstrating a separate preprocessing approach. 
Together with this, a custom Python script is made available, allowing researchers to generate various other configurations of preprocessing pipelines suited to their unique research needs (refer to Table \ref{tab:preprocessing} for the list of provided options).

%%%%%%%%%%%%%%%%%%%%%%%%%%%%%%%%%%%%%%%%%%%%%%%%%%%%%%%%%%%%%%%%
\begin{table}[tb!]
\footnotesize
\caption{Different preprocessing techniques for spatial, temporal, and amplitude normalization. Scripts are provided on the project's GitHub homepage to apply selected combinations of these techniques for specific needs.}
\begin{tabular}{@{}p{0.15\linewidth}p{0.8\linewidth}@{}}
\toprule
\textbf{Technique}         & \textbf{Description}                                                                                                 \\ \midrule
\textbf{Spatial} &                                                                                                                      \\
None                       & No spatial normalization.                                                                                             \\
Zero Padding               & Pad border of footstep with zeros to a specified tensor width and length.                                           \\
Resize                     & Apply spatial interpolation to resize the footstep to a specified tensor width and length.                           \\
Foot Rotation              & Rotate footstep according to the direction of its first principal component axis (the sole's longest dimension).     \\
Foot Translation           & Translate footstep according to the foot's center of area, mass, or bounding box centroid.     \\
Registration & Linearly transform (translate, rotate, scale) the footstep to align with a reference template (e.g., the MUN104 healthy barefoot templates \cite{Pataky2011}). \\
\textbf{Temporal}          &                                                                                                                      \\
None                       & No temporal normalization.                                                                                              \\
Zero Padding               & Pad the 3D tensor with zeros at the end of the footstep to a specified number of frames.                             \\
Interpolation  & Interpolate the 3D tensor to a specified number of frames (e.g., 101 frames, where each frame represents 1\% of the stance).                 \\
\textbf{Amplitude}         &                                                                                                                      \\
None                       & No amplitude normalization.                                                                                              \\
Body Mass                  & Linearly rescale amplitudes by the participant's measured body mass, so that amplitudes are relative to body weight. \\
Mean Pressure                   & Linearly rescale amplitudes by the average total pressure of the footstep.                                    \\
Min-Max                    & Linearly rescale amplitudes to a maximum value of 1 and minimum of 0.                                                \\
\bottomrule
\end{tabular}
\label{tab:preprocessing}
\vspace{-5mm}
\end{table}

\begin{itemize}[noitemsep,topsep=0pt]
\item \textit{Pipeline 1}: The first pipeline includes four steps: (1) spatial rotation, (2) spatial zero padding, (3) spatial translation, and (4) temporal interpolation. 
The footsteps were rotated according to the direction of their first principal component (PC) axis and flipped upright based on their direction of walking.
They were then spatially zero-padded to dimensions of $75 \times 40$ pixels, and translated to align their bounding box's centroid to the center of the padded area.
Lastly, nearest-neighbour interpolation was used to standardize all footsteps to 101 frames, where each frame represents a percentage of the footstep duration. 
This version of the preprocessed footstep data retains information such as foot size and body weight. 
However, information such as foot rotation angle and footstep duration has been uncoupled from the footstep recordings and can be retrieved from the metadata fields labeled as \textit{RotationAngle}, \textit{StartFrame}, and \textit{EndFrame}.
 
\item \textit{Pipeline 2}: The second pipeline builds upon the first by incorporating two additional components: (5) spatial resizing and (6) amplitude normalization using the mean total pressure. 
To normalize differences in foot size across participants, the sole dimensions were resized to a common size of 70 pixels in length and 25 pixels in width, corresponding to a size of $35 \times 12.5$ cm for the original sensor resolution. 
The original dimensions of the sole, representing the length and width of the foot's contact area with the sensor grid, were determined from the aligned footprints by counting the maximum number of active sensors (those with pressure exceeding 10 kPa) present along the two aligned axes during the footstep. 
To achieve the required dimensions, the pressure maps were resized via nearest neighbour interpolation, and samples were spatially zero-padded to $75 \times 40$ pixels for consistency with the other version of the footsteps. 
To normalize amplitude, the sensor data for each step was subsequently rescaled by dividing by the time-averaged sum of all sensor readings (i.e., the average total pressure). 
Compared to the initial pipeline, this approach further separates information related to foot size and body weight from the footstep measurements. 
The initial measurements for sole width, sole length, and average pressure are accessible in the \textit{Length}, \textit{Width}, and \textit{MeanPressure} metadata fields.
\end{itemize}

%%%%%%%%%%%%%%%%%%%%%%%%%%%%%%%%%%%%%%%%%%%%%%%%%%%%%%%%%%%%%%%%%%%%%%%%%%%%%%%%%%%
\subsubsection*{Labeling}\label{section:labeling}

The process of data labeling (i.e. data annotation) involves assigning target characteristics to training data, enabling statistical analysis or use in machine learning model development. 
In this dataset, labels were derived algorithmically from collected footsteps and subsequently confirmed through manual inspection (see details in the Manual Data Inspection section).

\begin{itemize}[noitemsep,topsep=0pt]
\item \textit{Side}: The left and right foot labels were determined through a pixel counting method outlined by MacDonald et al. \cite{Macdonald2023}. 
This method involves counting the number of activated sensors beneath different regions of the foot.
\item \textit{Orientation}: The orientation of each footstep, corresponding to the participant's direction of walking during each pass across the tiles, was determined by analyzing the COP trajectory of the foot (see Equation ({\ref{eq:COP}}) in the Technical Validation section).  
To determine the walking direction, the position of the anteroposterior (AP) COP in the beginning of the footstep was compared with its position at the end. 
A value of 1 indicates walking from Tiles 1 and 2 towards Tiles 11 and 12, while 0 indicates the opposite.
\item \textit{Incomplete}: Incomplete footsteps were defined as footsteps that fell partially outside of the tile grid or outside of the 90-second recording. 
These were detected by flagging footsteps with a starting time of $t = 0$ or an end time of $t = 90$, as well as footsteps that fell near the boundaries of the sensor platform and had a small area. 
Specifically, footsteps with a total number of activated sensors smaller than three scaled median absolute deviations (MAD) below the median footstep in the trial were flagged.
\item \textit{Standing}: Standing footsteps, observed during the slow-to-stop trials, were also identified by analysis of the COP trajectory of the foot. 
Linear least-squares regression was used to estimate the slope of the AP COP for each footstep, and footsteps with a slope exceeding three MAD from the median footstep in the trial were identified as standing footsteps.
\end{itemize}

%%%%%%%%%%%%%%%%%%%%%%%%%%%%%%%%%%%%%%%%%%%%%%%%%%%%%%%%%%%%%%%%%%%%%%%%%%%%%%%%%%%
%%%%%%%%%%%%%%%%%%%%%%%%%%%%%%%%%%%%%%%%%%%%%%%%%%%%%%%%%%%%%%%%%%%%%%%%%%%%%%%%%%%
%%%%%%%%%%%%%%%%%%%%%%%%%%%%%%%%%%%%%%%%%%%%%%%%%%%%%%%%%%%%%%%%%%%%%%%%%%%%%%%%%%%
\section*{Data Records}
% The Data Records section should be used to explain each data record associated with this work, including the repository where this information is stored, and to provide an overview of the data files and their formats. Each external data record should be cited numerically in the text of this section, for example \cite{Hao:gidmaps:2014}, and included in the main reference list as described below. A data citation should also be placed in the subsection of the Methods containing the data-collection or analytical procedure(s) used to derive the corresponding record. Providing a direct link to the dataset may also be helpful to readers (\hyperlink{https://doi.org/10.6084/m9.figshare.853801}{https://doi.org/10.6084/m9.figshare.853801}).

%Tables should be used to support the data records, and should clearly indicate the samples and subjects (study inputs), their provenance, and the experimental manipulations performed on each (please see 'Tables' below). They should also specify the data output resulting from each data-collection or analytical step, should these form part of the archived record.

%%%%%%%%%%%%%%%%%%%%%%%%%%%%%%%%%%%%%%%%%%%%%%%%%%%%%%%%%%%%%%%%%%%%%%%%%%%%%%%%%%%
\begin{figure}[b!]
\footnotesize
%\centering % gave an error, would not compile  
    \begin{forest}
        for tree={
            font=\sffamily, grow'=0,
            folder indent=.9em, folder icons,
            edge= dotted
        }
        [StepUP-P150
          [participant\_metadata.csv, is file]
          [\{py{,}mat\}
          [001 
              [BF
                [S1
                    [trial.\{npz{,}mat\},is file]
                    [preprocessed.\{npz{,}mat\},is file]
                ]
                [S2]
                [S3]
                [W1
                    [metadata.csv, is file]
                    [trial.\{npz{,}mat\}, is file]
                    [pipeline\_1.\{npz{,}mat\}, is file]
                    [pipeline\_2.\{npz{,}mat\}, is file]
                ]
                [W2]
                [W3]
                [W4]
              ]
              [ST]
              [P1]
              [P2]
              ]
          [002]
          [003]
          [\\\vdots, is file]
          [150]
        ]
        ]
    \end{forest}
    \caption{Data structure for the UNB StepUP-P150 dataset.}\label{fig:tree_directory}
\end{figure}
%%%%%%%%%%%%%%%%%%%%%%%%%%%%%%%%%%%%%%%%%%%%%%%%%%%%%%%%%%%%%%%%%%%%%%%%%%%%%%%%%%%

The StepUP-P150 dataset is accessible for download on the \emph{Federated Research Data Repository (FRDR)} \cite{UNB-StepUP} (\url{https://doi.org/10.20383/103.01285}) and contains the raw recordings, two variants of extracted and preprocessed footstep data, and detailed metadata for each individual footstep. 
The files are organized as shown in Fig. \ref{fig:tree_directory}. 
Located at the top level, the spreadsheet named `participant\_metadata.csv' contains demographic information, anthropometric measurements, types of footwear, and other possible influences on the gait patterns of each for the 150 participants. 
More comprehensive information regarding the metadata fields is available in Table \ref{tab:participant_metadata}. 
For convenience, both Python-compatible and MATLAB-compatible versions of the footstep data are provided, organized into the top-level folders `py' and `mat', respectively.
For each version, the pressure recordings are structured into 150 folders, with each corresponding to a single participant.
Each folder is named using the pattern XXX, with `XXX' representing a unique, randomly assigned ID that encompasses all participants (001-150). 
Within each participant's folder, four (third-level) sub-folders represent the various footwear conditions: `BF', `ST', `P1', and `P2'. 
Inside these, there are seven further (fourth-level) sub-subfolders designated for distinct standing balance and walking experiment trials: `S1', `S2', `S3', `W1', `W2', `W3', or `W4' (refer to Fig. \ref{fig:protocol_tasks} and the Experimental Protocol section for more information). 

%%%%%%%%%%%%%%%%%%%%%%%%%%%%%%%%%%%%%%%%%%%%%%%%%%%%%%%%%%%%%%%%%%%%%%%%%%

\begin{table}[b!]
\caption{Metadata fields, possible values/format, and description for the spreadsheet `participant\_metadata.csv' containing demographic information, anthropometric measurements, types of footwear, and other possible influences on the gait patterns of each of the 150 participants.}
\footnotesize
\begin{tabular}{@{}p{0.27\linewidth}p{0.24\linewidth}p{0.44\linewidth}@{}}
\toprule
\textbf{Field}                  & \textbf{Possible Values/Format} & \textbf{Description}                                                       \\ \midrule
ParticipantID                   & 001-150                         & Unique participant identifier.                                                         \\
Sex                             & Male, Female                    & Sex assigned at birth.                                                     \\
Gender                          & Man, Woman, Non-Binary, Not Specified & Gender identity.                                                           \\
Age                             & 19-91                           & Age in years.                                                              \\
RaceEthnicity & Black, East/Southeast Asian, Middle Eastern, South Asian, White, Other, Unknown            & Population group (race or ethnicity).           \\
RaceEthnicityOther              & Text                            & Description of race/ethnicity if selected `Other'.                         \\
DominantLeg                     & Left, Right                     & Self-reported dominant side for a kicking task.                            \\
Weight (Kg)                     & 46.3-148.4                      & Measured weight in Kg at time of collection.                               \\
Height (Kg)                     & 151.0-195.5                     & Measured height in cm at time of collection.                               \\
LeftFootLength (cm)             & 21.5-30.0                       & Measured left foot length in cm.                                           \\
LeftFootWidth (cm)              & 7.0-11.0                        & Measured left foot width in cm.                                            \\
RightFootLength (cm)            & 20.0-30.0                       & Measured right foot length in cm.                                          \\
RightFootWidth (cm)             & 7.5-11.0                        & Measured right foot width in cm.                                           \\
StandardShoeSize                & 4-12.5                          & Chosen standard shoe size for the ST trials (UK sizing).                   \\
Shoe1Category & Athletic, Boots, Business/Dress, Casual Sneaker, Flat Canvas, Hiking/Trail, Sandals, Other & Category of participant's first personal shoe (P1). \\
Shoe1Size                       & Varied                          & Shoe size for P1, in varied units (e.g., US M, US W, EU, UK).              \\
Shoe1Brand                      & Text                            & Shoe brand name for P1.                                                    \\
Shoe1Description                & Text                            & Additional description or detail about P1.                                 \\
Shoe2Category                   & See Shoe1Category               & Repeated fields for participant's second shoe (P2).                        \\
Shoe2Size                       & See Shoe1Size                   &                                                                            \\
Shoe2Brand                      & See Shoe1Brand                  &                                                                            \\
Shoe2Description                & See Shoe1Description            &                                                                            \\
BFType                          & Barefoot, Socks                 & Whether BF trials were performed barefoot or wearing socks.                \\
OrthopedicInjury                & Yes, No                         & Recent orthopedic injury or surgery (e.g., hip, knee).                     \\
OrthopedicInjuryComment         & Text                            & Comment if selected `Yes' for OrthopedicInjury.                            \\
AssistiveDevice                 & Yes, No                         & Regular use of assistive device (e.g., cane, walker).                      \\
AssistiveDeviceComment          & Text                            & Comment if selected `Yes' for AssistiveDevice.                             \\
MusculoskeletalCondition        & Yes, No                         & Musculoskeletal condition (e.g., arthritis).                               \\
MusculoskeletalConditionComment & Text                            & Comment if selected `Yes' for MusculoskeletalCondition.                    \\
RecentExercise                  & Yes, No                         & Muscle pain or soreness due to recent exercise or other activity.          \\
RecentExerciseComment           & Text                            & Comment if selected `Yes' for RecentExercise.                              \\
OtherCondition                  & Text                            & Additional comments on other conditions that may impact walking behaviour. \\ \bottomrule
\end{tabular}
\label{tab:participant_metadata}
\end{table}

%%%%%%%%%%%%%%%%%%%%%%%%%%%%%%%%%%%%%%%%%%%%%%%%%%%%%%%%%%%%%%%%%%%%%%%%%%

For each of the three standing balance trials (S1, S2, and S3), there are two corresponding files. 
The `trial.\{npz,mat\}' file houses the raw, unprocessed 30-second recording from the trial, represented as a 3D tensor with dimensions of nominally $3000 \times 720 \times 240$ (frames, height, width; note that the number of frames may vary slightly). 
The processed version of the recording is available as `preprocessed.\{npz,mat\}', representing a 3D tensor with dimensions of $3000 \times 180 \times 180$ (frames, height, width). 
As noted, these files are available in both NumPy (.npz) and MATLAB (.mat) file formats, each of which uses a dictionary-like structure for variable storage, with the tensors accessible under the top-level key arr\_0.

Unlike the balance trials, each folder for the walking trials (W1, W2, W3, and W4) includes four files. 
The raw, unprocessed 90-second pressure data is available in `trial.\{npz,mat\}', containing a 3D tensor with dimensions of $9000 \times 720 \times 240$ (frames, height, width; again, the number of frames may vary). 
The footsteps were extracted using two distinct preprocessing methods and are provided as `pipeline\_1.\{npz,mat\}' and `pipeline\_2.\{npz,mat\}', which are explained in the Footstep Normalization section. 
Each tensor is 4D with dimensions $n_{footsteps} \times 101 \times 75 \times 40$ (samples, frames, height, width), where $n_{footsteps}$ specifies the number of footsteps in a given trial. 
These tensors are all accessible via the key arr\_0. 
Each individual footstep's metadata is provided in a file named `metadata.csv', containing details such as the footstep's spatiotemporal location in the initial recording, classification labels (e.g., left or right foot, walking direction, outlier), and various parameters obtained during preprocessing (e.g., rotation angle, length and width of the foot's contact area, mean pressure). 
Table \ref{tab:metadata} lists the metadata fields and their explanations.

\begin{table}[b!]
\begin{threeparttable}[b]
\caption{Metadata fields, possible values/format, and description for the `metadata.csv' spreadsheets containing information regarding trial conditions, 3D bounding boxes for each footstep, footstep labels, and other parameters extracted during data processing. There is one `metadata.csv' file for each 90-second trial recording.}
\label{tab:metadata}
\footnotesize
\begin{tabular}{@{}p{0.15\linewidth}p{0.20\linewidth}p{0.55\linewidth}@{}}
\toprule
\textbf{Field} &
  \textbf{Possible Values/Format} &
  \textbf{Description} \\ \midrule
ParticipantID &
  001-150 &
  Unique participant identifier. \\
Footwear &
  BF, ST, P1, P2 &
  Footwear condition for trial. \\
Speed &
  W1, W2, W3, W4 &
  Walking speed condition for trial. \\
FootstepID\tnote{*} &
  0-$N_{steps}$ &
  Footstep's index in 90-second trial. \\
PassID\tnote{*} &
  0-$N_{passes}$ &
  The pass within which the footstep occurred (i.e., incremented each time the participant steps off the tiles to turn around). \\
StartFrame\tnote{*} &
  0-$N_{frames}$ &
  Frame in 90-second recording where footstep's first pressure contact occurred. \\
EndFrame\tnote{*} &
  0-$N_{frames}$ &
  Frame in 90-second recording where footstep's pressure contact ended. \\
Ymin, Ymax\tnote{*} &
  0-719 &
  Footstep bounding box limits along tile grid $y$-axis (parallel to walking direction, 3.6 m length). \\
Xmin, Xmax\tnote{*} &
  0-239 &
  Footstep bounding box limits along tile grid $x$-axis (perpendicular to walking direction, 1.2 m width). \\
Orientation &
  0, 1 &
  Footstep's orientation on the tile grid: 1 if walking toward Tiles 11 and 12, or 0 if walking toward Tiles 1 and 2.\\
Side &
  Left, Right &
  Whether the footstep corresponds to the right or left foot. \\
Standing &
  0, 1 &
  1 the footstep corresponds to standing behaviour during the Slow-to-Stop (W2) trials, 0 otherwise. \\
Incomplete &
  0, 1 &
  1 if the footstep was not captured fully by the sensors (e.g., partially off of the tile-grid or cut-off at beginning or end of recording), 0 otherwise. \\
Rscore &
  0-89 &
  Footstep's R-score, approximating the number of standard deviations from the trial mean. \\
Outlier &
  0, 1 &
  1 if the footstep's R-Score exceeds the recommended threshold of 2, or 0 otherwise. \\ 
Exclude &
  0, 1 &
  The combination of the `Standing', `Incomplete' and `Outlier' columns for easy exclusion of these footsteps if desired. \\ 
RotationAngle &
  -90\textdegree - 90\textdegree &
  Footstep's rotation angle in degrees with respect to the tile grid's long axis ($y$-axis). \\
FootLength &
  1-75 &
  Length of the sole's contact area in pixels, measured along the footstep's first principal component axis (longest foot dimension). \\
FootWidth &
  1-40 &
  Width of the sole's contact area in pixels, measured along the footstep's second principal component axis (perpendicular to longest foot dimension). \\
MeanPressure &
  Floating Point &
  Average of the footstep's total pressure (sum of all pressure values in kPa at each time point) over the duration of the stance. \\
  \bottomrule
\end{tabular}
\begin{tablenotes}
       \item [*] zero-indexed value
\end{tablenotes}
\end{threeparttable}
\end{table}

%%%%%%%%%%%%%%%%%%%%%%%%%%%%%%%%%%%%%%%%%%%%%%%%%%%%%%%%%%%%%%%%%%%%%%%%%%%%%%%%%%%

%%%%%%%%%%%%%%%%%%%%%%%%%%%%%%%%%%%%%%%%%%%%%%%%%%%%%%%%%%%%%%%%%%%%%%%%%%%%%%%%%%%
\section*{Technical Validation}

% This section presents any experiments or analyses that are needed to support the technical quality of the dataset. This section may be supported by figures and tables, as needed. This is a required section; authors must present information justifying the reliability of their data.

%%%%%%%%%%%%%%%%%%%%%%%%%%%%%%%%%%%%%%%%%%%%%%%%%%%%%%%%%%%%%%%%%%%%%%%%%%%%%%%%%%%
\subsection*{Data Quality Assessment} % Data Inspection / Statistical Analysis/Biometric Quality Assessment

\subsubsection*{Manual Data Inspection}

\begin{figure}[!t]
\centering
    \includegraphics[width=0.85\textwidth]{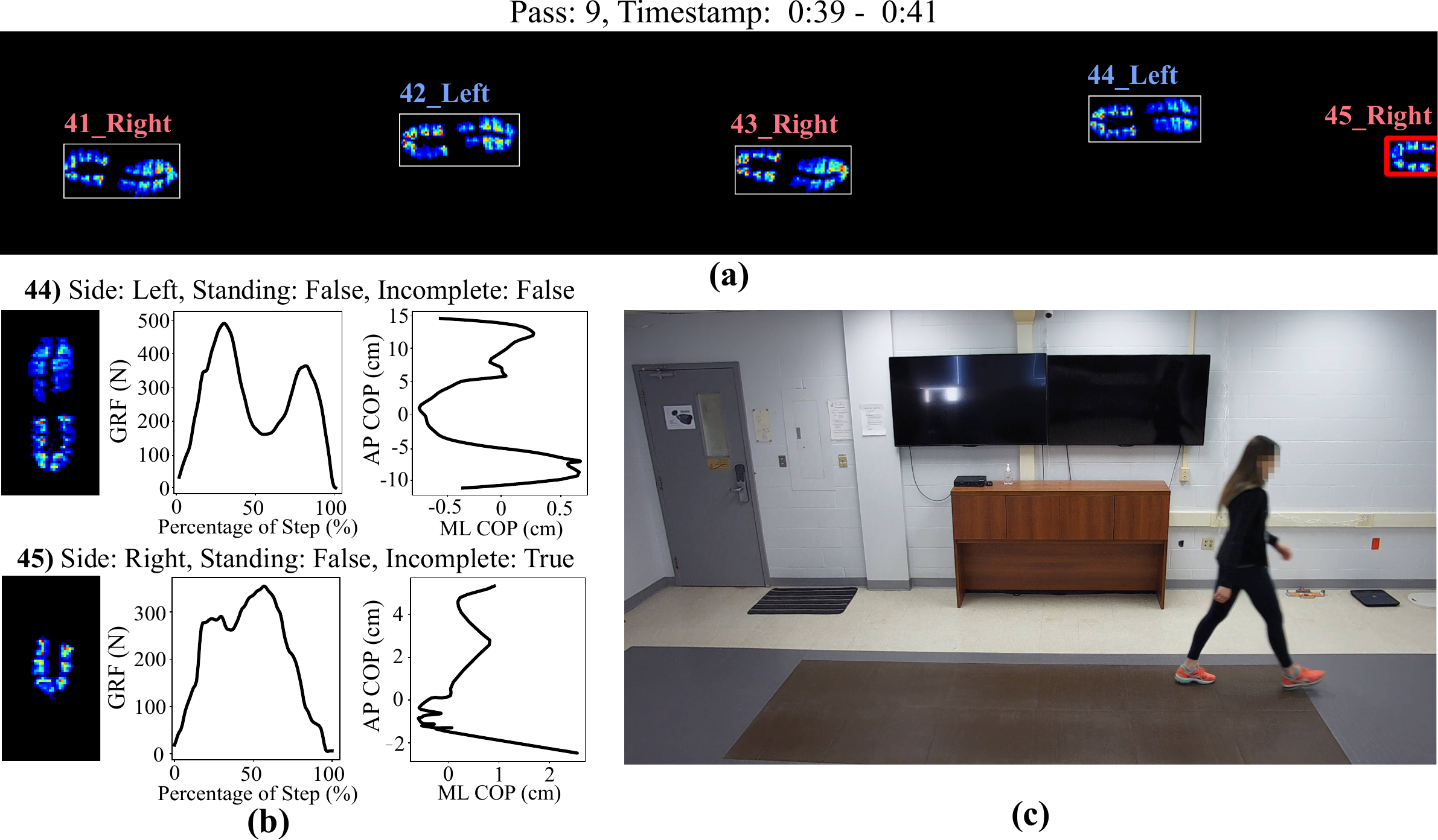}
    \caption{Examples of different views of the data used for manual inspection of footstep bounding boxes and labels in each recorded trial; (a) a ``multiple footstep'' view, which depicts the peak pressures for each pass across the tiles along with the detected footstep bounding boxes, left/right labels, and whether the footstep was flagged as an incomplete or standing footstep (e.g., in this example, footstep 45 is color-coded in red to indicate an incomplete footstep), (b) an ``individual footstep'' view that shows the peak pressure image, GRF time series, and COP trajectory for each footstep along with the predicted footstep labels, and (c) a frame of the corresponding RGB video from Camera 7.}
    \label{fig:manual_inspection}
\vspace{-4mm}
\end{figure}

As outlined in the prior section on Data Processing, metadata labels were identified through automated algorithms. 
Although these algorithms reached impressive accuracy levels (such as $99.7\%$ for identifying left and right footsteps \cite{Macdonald2023} for the \textit{Side} metadata field), the necessity for perfectly accurate labels in benchmark datasets is crucial. 
Therefore, a minimum of two research team members performed a manual/visual inspection of each footstep and its associated metadata within the dataset. 
A multi-stage manual inspection of the extracted footsteps was conducted, where any errors identified in the initial review were addressed before starting the second review by a different evaluator. 
The evaluators conducted assessments of each footstep by visually inspecting the raw data for each pass across the tiles (Fig. \ref{fig:manual_inspection}(a)), along with processed versions of the data for each segmented step (Fig. \ref{fig:manual_inspection}(b)). 
This included analyzing gait features such as aligned foot peak pressure images, vertical ground reaction force (GRF) profiles, foot center of pressure (COP) trajectories (Fig. \ref{fig:manual_inspection}(b)), as well as reviewing associated video recordings (captured by Camera 7, positioned at a 90\textdegree~viewing angle, perpendicular to the main axis of the tile grid; Fig. \ref{fig:manual_inspection}(c)). 
These inspection phases were also used to identify and eliminate any overlooked data corruption, hardware malfunctions, or protocol complications during data collection.

%%%%%%%%%%%%%%%%%%%%%%%%%%%%%%%%%%%%%%%%%%%%%%%%%%%%%%%%%%%%%%%%%%%%%%%%%%%%%%%%%%%

Specifically, the assessors recorded the errors in spreadsheets designated for each trial and merged the necessary corrections into the dataset where applicable. 
Errors fell into two primary categories: those concerning bounding boxes and those related to metadata. 
An interactive tool was developed to enable evaluators to manually adjust the size of spatiotemporal bounding boxes with inaccurate dimensions. 
Common bounding box errors included (1) bounding boxes that were too small spatiotemporally to contain a complete footstep, often occurring with high-arch shoes where the algorithm identified just a portion of the pressure profile, such as the heel or toe area; (2) bounding boxes that were too large, encroaching on the spatiotemporal area of another footstep, commonly observed in the W2 trials when participants slightly shuffled their feet during the slow-to-stop walking maneuver; and (3) missing boxes, especially for footsteps that occurred mostly outside the tile grid (e.g., a footstep that fell only 20\% within the grid, deemed an incomplete footstep).
When common errors were noted in the auto-labeling, corrections were implemented programmatically whenever feasible. 
Errors most commonly occurred (1) in the \textit{Side} metadata field, where a left label was incorrectly matched with a right footstep or vice versa, especially for incomplete footsteps; (2) in the \textit{Standing} metadata field, where certain walking behaviours, such as shuffling during a slow-to-stop maneuver, resulted in inconsistent label assignment by the algorithm; and (3) in the \textit{Incomplete} metadata field, where partial footsteps were not flagged as incomplete steps.

%%%%%%%%%%%%%%%%%%%%%%%%%%%%%%%%%%%%%%%%%%%%%%%%%%%%%%%%%%%%%%%%%%%%%%%%%%%%%%%%%%%
\subsubsection*{Automated Outlier Detection}

\begin{figure}[tb!]
    \centering
    \includegraphics[width=0.6\linewidth]{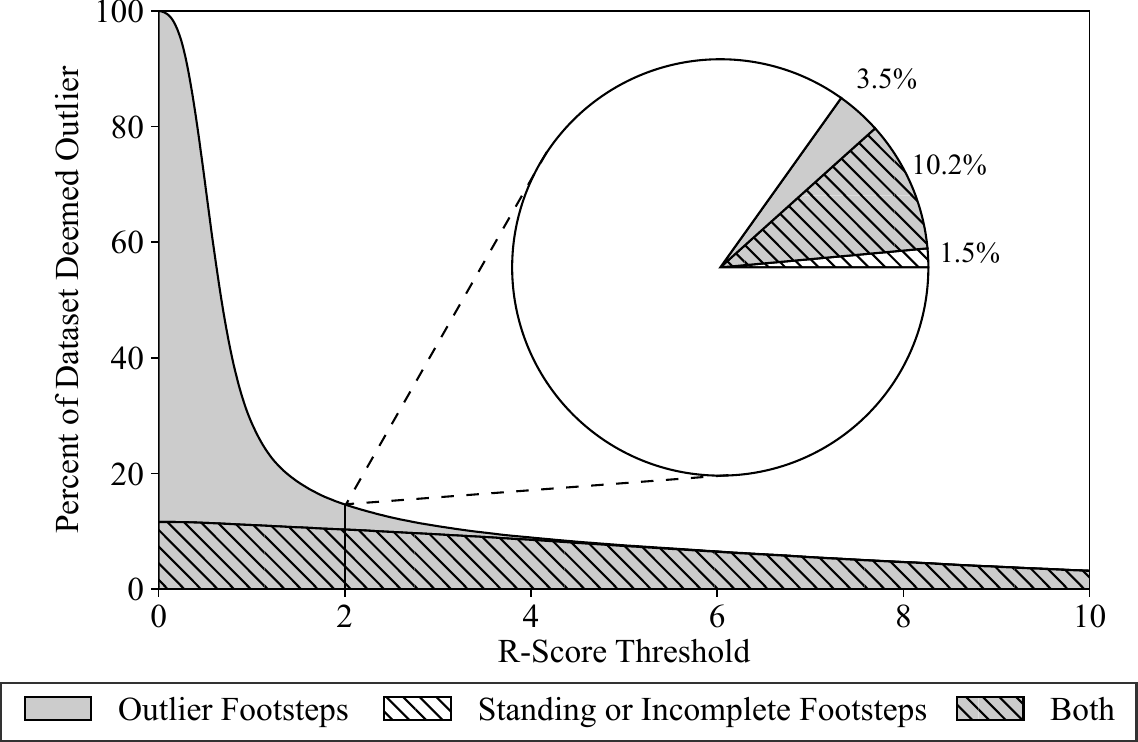}
    \caption{Performance of R-score for detecting outliers across cutoff thresholds between 0.0 and 10.0. Note: The shaded area represents the percentage of samples marked as outliers at a specific R-score threshold, with a hatched pattern indicating standing or incomplete footsteps (detected during labeling and manual inspection). In this study, a score threshold of 2 was chosen, leading to 13.7\% of the footsteps being categorized as outliers, with 3.5\% being regular steps that had not been identified in previous manual inspection.}
    \label{fig:outlier}
\vspace{-5mm}
\end{figure}

To enable a focus on only high-quality footsteps, a metadata field titled \textit{Outlier} was introduced, complementing the existing \textit{Incomplete} metadata field for partial or incomplete footsteps, allowing researchers to omit outlier samples. 
Certain footsteps may be of lesser quality for gait recognition or gait analysis due to irregular gait patterns such as stumbling or changes in speed or direction, as well as factors related to hardware or sensors, such as ghosting (i.e., some sensors remaining activated after a footstep), disconnected sensors, or noise. 
To identify high-quality, representative footsteps and exclude potentially inferior ones, an approach based on the R-score described by Sangeux and Polak \cite{SangeuxPolak2015} was used. 
Specifically, all of the footsteps in a trial were compared to a representative, median footstep for that trial, and their similarity (and by association, quality) was quantified by an R-score. 
For normally distributed measurements, the R-score approximately represents the number of standard deviations from the mean. 
The scores were calculated using the combination of a spatial characteristic (i.e., the number of active sensors during the footstep), a temporal characteristic (i.e., the duration of the footstep), and an amplitude characteristic (i.e., the footstep's GRF profile). 
The R-scores were independently computed for each 90-second trial because variations in footwear and walking speed are likely to affect the normality assumption of this method. 
Moreover, the representative median was calculated without manually-identified standing or incomplete footsteps.
Footsteps within the dataset with an R-score of 2.0 or higher were marked as outliers, labeled \textit{Outlier} in the metadata. 
This averages to about 12.3 footsteps per trial, culminating in a total of 29,511 flagged footsteps throughout the entire dataset. 
It should be noted that many of these footsteps had already been classified as \textit{Standing} or \textit{Incomplete} (Fig. \ref{fig:outlier}). 
Metadata in the form of \textit{Rscore} is also included, allowing researchers to establish their own exclusion criteria if desired.

%%%%%%%%%%%%%%%%%%%%%%%%%%%%%%%%%%%%%%%%%%%%%%%%%%%%%%%%%%%%%%%%%%%%%%%%%%%%%%%%%%%
%%%%%%%%%%%%%%%%%%%%%%%%%%%%%%%%%%%%%%%%%%%%%%%%%%%%%%%%%%%%%%%%%%%%%%%%%%%%%%%%%%%

\subsection*{Comparison to Existing Datasets}
The dataset's technical quality and reliability were evaluated by analyzing different common gait representations and parameters. 
This evaluation was conducted in comparison to other publicly available datasets that use pressure or force measurements for gait.

\subsubsection*{Spatial Representations}
\begin{figure}[tb!]
    \centering
    \includegraphics[width=0.7\linewidth]{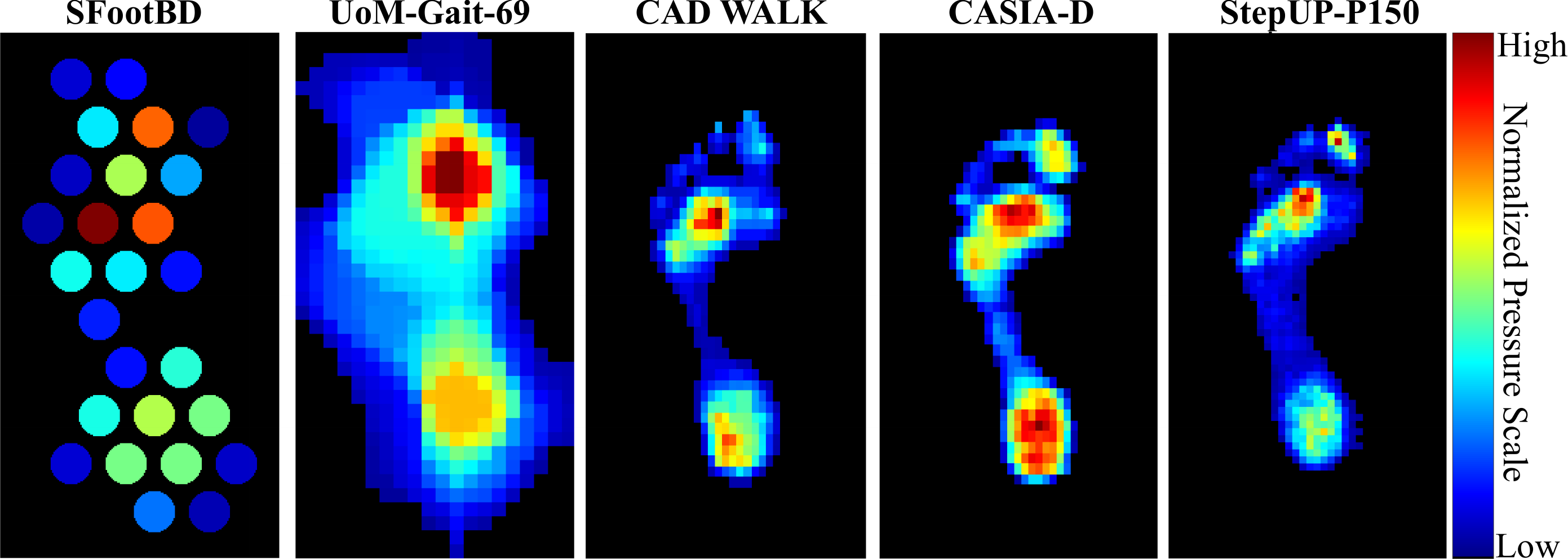}
    \caption{Sensor resolution (or sensor density) comparison of floor-sensing platforms for pressure-based gait databases (SFootBD, UoM-Gait-69, CAD WALK, CASIA-D, and UNB StepUP-P150). Note: The SFootBD dataset used a custom system consisting of piezoelectric sensors with diameters of 27 mm; the UoM-Gait-69 dataset used a custom system (iMAGiMAT) consisting of 116 plastic optical fibres (POFs), from which a spatial reconstruction was estimated using the Landweber algorithm; the CAD WALK and CASIA-D datasets used RS Scan Footscan platforms with a spatial resolution of 7.62 mm $\times$ 5.08 mm; and the UNB StepUP-150 dataset used Stepscan tiles with a spatial resolution of 5 mm $\times$ 5 mm.   
    %In this figure, all peak pressure images have been resized to common dimensions for spatial comparison (1 pixel $=$ 1 mm $\times$ 1 mm).
    }
    \label{fig:resolution_comparison}
\vspace{-5mm}
\end{figure}

Within the literature on plantar pressure, it is typical to employ a range of two-dimensional spatial representations to distill specific characteristics of each pixel's values over time, such as its peak pressure, pressure-time integral, contact duration, and time-to-maximum \cite{Pataky2012}. 
This approach synthesizes all pertinent details from the time series of pressure images throughout a stance phase into a comprehensive image. 
Among these features, peak pressure images are probably the most widely employed in the literature, and are computed from each footstep's 3D tensor as the maximum pressure experienced by each pixel (i.e., sensor) during the stance. 
Peak pressure images were used here to illustrate the variations in unshod and shod pressure patterns (Fig. \ref{fig:P100s}) as well as the differences in sensor density across datasets (Fig. \ref{fig:resolution_comparison}). 

Compared to other currently available pressure-based gait datasets, StepUP-P150 offers the highest spatial resolution at $5 \times 5$ mm (Fig. {\ref{fig:resolution_comparison}}). 
This represents a more than 50\% increase along the axis of walking compared to the next two highest-resolution underfoot pressure databases: CAD WALK and CASIA-D. 
The peak pressure images in Fig.~{\ref{fig:P100s}} illustrate the high quality of texture and shape information captured in the footstep recordings. 
Distinct differences in pressure distributions are visually apparent across participants, even when wearing the same shoe type, highlighting the individuality of gait patterns. 
Key outsole characteristics and high-pressure contact regions are also clearly visible, revealing detailed features of the participants' personal footwear.
%A much higher sensing resolution translates to a significantly larger number of voxels per footstep. 
%Until now, high-resolution underfoot pressure data was only available in much smaller datasets, such as the CASIA-D dataset, which is arguably the most well-known in this field and contains just over 3,000 footsteps from 118 individuals \cite{Zheng2011}.

%%%%%%%%%%%%%%%%%%%%%%%%%%%%%%%%%%%%%%%%%%%%%%%%%%%%%%%%%%%%%%%%%%%%%%%%%%%%%%%%%%%
%%%%%%%%%%%%%%%%%%%%%%%%%%%%%%%%%%%%%%%%%%%%%%%%%%%%%%%%%%%%%%%%%%%%%%%%%%%%%%%%%%%

\subsubsection*{Time Series Representations}
Biomechanical analyses also frequently involve time series signals due to the temporal nature of gait patterns, such as vertical ground reaction force (GRF) profiles, and center of pressure (COP) trajectories.
GRF time series represent the forces exerted by the ground on the foot throughout the stance phase, and are computed by summing the pressures ($p$) across all pixels at each time step and converting to force in Newtons (N) using the area of each sensor ($A = 2.5 \times 10^{-5} \mathrm{m}^2$), as in ({\ref{eq:GRF}}).
\begin{align}
    \mathrm{GRF}(t)\mathrm{~[N]}= 1000\times A \mathrm{~[m^2]} \times \sum_{x,y} p_{xy}(t) \mathrm{~[kPa]}
    \label{eq:GRF}
\end{align}
COP time series are defined in the mediolateral (ML, $x$-axis) and anteroposterior (AP, $y$-axis) directions, and are calculated as the pressure-weighted average of the foot's coordinates at each time step during the stance, as in ({\ref{eq:COP}}) where $s$ represents the sensor side length ($s = 0.5 \mathrm{~cm}$).
\begin{align}
    \mathrm{COP_{ML}}(t)\mathrm{~[cm]}= \frac{\sum_{x,y} x \times p_{xy}(t)}{\sum_{x,y} p_{xy}(t)}\times s \mathrm{~[cm]}~~~~~~~~~~
    \mathrm{COP_{AP}}(t)\mathrm{~[cm]}= \frac{\sum_{x,y} y \times p_{xy}(t)}{\sum_{x,y} p_{xy}(t)}\times s \mathrm{~[cm]}
    \label{eq:COP}
\end{align}
The COP time series were further mean-centered to remove absolute position offsets and emphasize relative shifts in pressure distribution throughout each step.

\begin{figure}[tb!]
\centering
    \includegraphics[width=0.8\textwidth]{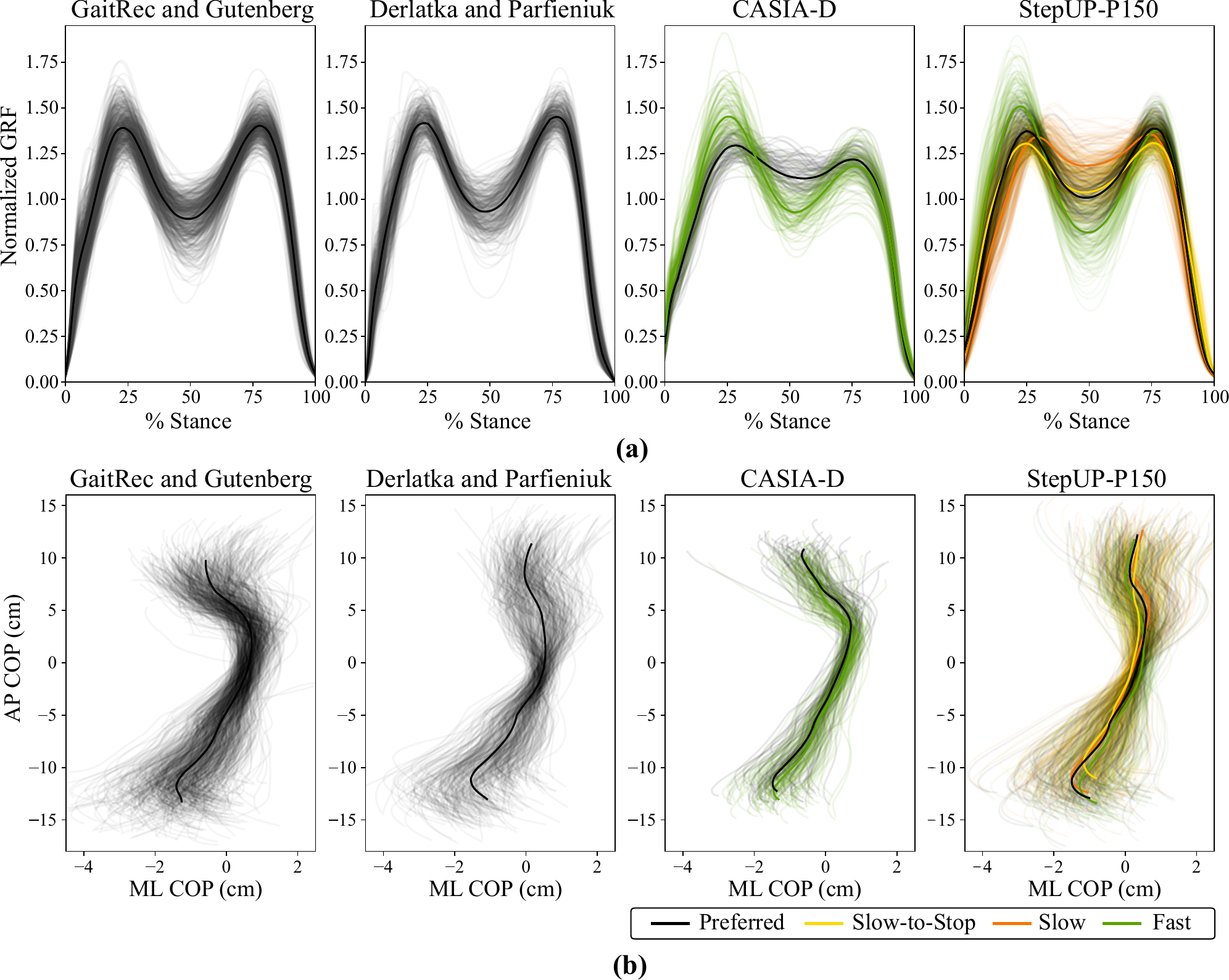}

    \caption{Comparison of average ground reaction force (GRF) and foot center of pressure (COP) time series from UNB StepUP-P150 and from three public gait datasets: (1) healthy participants from the GaitRec and Gutenberg databases (Kistler force plate measurements from 561 individuals walking mostly barefoot), (2) the Dertlaka and Parfieniuk database (Kistler force plate measurements from 324 individuals walking in shoes), and (3) the CASIA-D barefoot database (RS Scan Footscan pressure measurements from 88 individuals that walked barefoot at two speeds). Note: Each GRF time series was rescaled by its mean value for ease of comparison across datasets.}
    \label{fig:GRFCOP_comparison}
\vspace{-5mm}
\end{figure}

The GRF and COP time series were used to assess data quality and compare StepUP-P150 with publicly available datasets from healthy individuals (Fig.~{\ref{fig:GRFCOP_comparison}}). 
These included two datasets that used force plates (GaitRec and Gutenberg {\cite{Horsak2020,Horst2021}}; Derlatka and Parfieniuk {\cite{DerlatkaParfieniuk2023}}) and one that used a pressure sensing walkway (CASIA-D barefoot {\cite{Zheng2011}}).
For visualization, the StepUP-P150 signals were averaged across all footwear conditions and grouped by the four walking speeds.
The resulting average GRF and COP waveforms are comparable to those from existing datasets, especially the Derlatka and Parfieniuk dataset which also included participants walking in footwear.
The other two datasets primarily or exclusively included barefoot walking.
Variation across walking speeds is also evident in the StepUP-P150 data, such as an increase in peak GRF during heel strike for faster walking.
This is consistent with the trends observed for the CASIA-D dataset and prior biomechanics studies {\cite{Segal2004}}.
%By capturing gait across a range of walking speeds, StepUP-P150 offers increased variability and flexibility, providing broader research opportunities. 

%%%%%%%%%%%%%%%%%%%%%%%%%%%%%%%%%%%%%%%%%%%%%%%%%%%%%%%%%%%%%%%%%%%%%%%%%%%%%%%%%%%
%NOTE: we normally apply rotation during preprocessing which removes foot angle information  - here I kept the foot rotation before calculating COP. 
%%%%%%%%%%%%%%%%%%%%%%%%%%%%%%%%%%%%%%%%%%%%%%%%%%%%%%%%%%%%%%%%%%%%%%%%%%%%%%%%%%%

\subsubsection*{Spatiotemporal Gait Parameters}

\begin{figure}[tb!]
    \centering
    \includegraphics[width=0.7\linewidth]{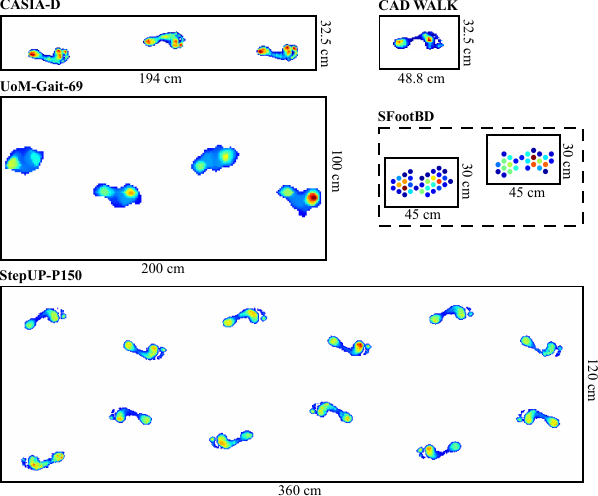}
    \caption{Size comparison of floor-sensing platforms for pressure-based gait databases (CASIA-D, CAD WALK, UoM-Gait-69, SFootBD, and UNB StepUP-P150)}
    \label{fig:sensor_comparison}
\vspace{-4mm}
\end{figure}

As shown in Fig.~{\ref{fig:sensor_comparison}}, the 120 cm $\times$ 360 cm sensing tile grid used for the StepUP-P150 dataset is substantially larger than those used in previous pressure-based gait datasets. 
This expanded area enables the capture of natural walking behavior across 4–6 consecutive footsteps, compared to 1–4 steps in prior datasets, and eliminates the need for subjects to target small, fixed sensing areas.
As a result, spatiotemporal gait parameters such as gait speed, cadence, step length, step width, and toe-out angle can be reliably extracted from the pressure recordings. 
These parameters were compared to those from the CASIA-D dataset, the only other currently available public dataset with sufficient spatial resolution and sequential step count to support such analysis.
Except for the toe-out angle, all features were computed from distances between 3D bounding box coordinates ($time$, $x$, $y$) of consecutive footsteps. 
The toe-out angle was calculated as the angle between the foot’s first principal component axis (i.e., its longest dimension) and the long axis of the tile grid, with negative values indicating inward rotation and positive values indicating outward rotation.

%%%%%%%%%%%%%%%%%%%%%%%%%%%%%%%%%%%%%%%%%%%%%%%%%%%%%%%%%%%%%%%%%%%%%%%%%%%%%%%%%%%
%%%%%%%%%%%%%%%%%%%%%%%%%%%%%%%%%%%%%%%%%%%%%%%%%%%%%%%%%%%%%%%%%%%%%%%%%%%%%%%%%%%

\begin{figure}[tb!]
    \centering
    \includegraphics[width=\linewidth]{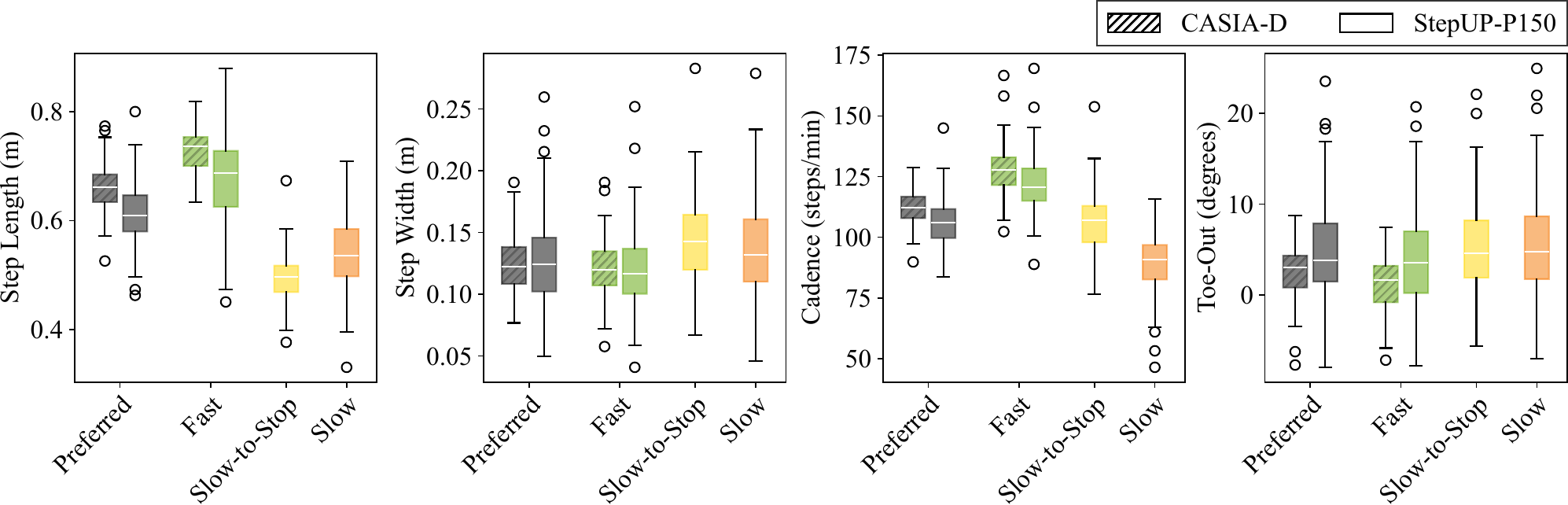}
    \caption{Comparison of four spatiotemporal gait parameters between the CASIA-D barefoot database and barefoot samples from the UNB StepUP-P150 dataset. Note: Significant differences were found between the CASIA-D and StepUP-P150 datasets in step length, cadence, and toe-out angle, for both the preferred and fast walking trials ($p < 0.05$ for all, using two-sample $t$-tests). No significant differences were found for step width.}
    \label{fig:spatiotemporal_boxplot}
\vspace{-4mm}
\end{figure}

Notably, significant differences in step length, cadence, and toe-out angle were observed between CASIA-D and StepUP-P150 (Fig.~{\ref{fig:spatiotemporal_boxplot}}, $p < 0.05$ using two-sample $t$-tests). 
These differences likely stem from sensor size and protocol variations: (1) the wider sensing platform for StepUP-P150 may have allowed for greater toe-out angles, and (2) participants turned and sometimes paused between passes, potentially resulting in slower walking speeds that reduced step length and cadence. 
Indeed, participants in the StepUP-P150 dataset walked 0.21 m/s slower on average than normative speeds, stratified by sex and age group {\cite{Bohannon2011}}.
However, the StepUP-P150 dataset also exhibited higher variability across the four gait parameters compared to CASIA-D.
The larger participant pool, multiple footwear conditions, and varied walking speeds in StepUP-P150 contribute to a broader and more representative sample of gait behavior.

%%%%%%%%%%%%%%%%%%%%%%%%%%%%%%%%%%%%%%%%%%%%%%%%%%%%%%%%%%%%%%%%%%%%%%%%%%%%%%%%%%%

%%%%%%%%%%%%%%%%%%%%%%%%%%%%%%%%%%%%%%%%%%%%%%%%%%%%%%%%%%%%%%%%%%%%%%%%%%%%%%%%%%%

\begin{figure}[tb!]
    \centering
    \includegraphics[width=\textwidth]{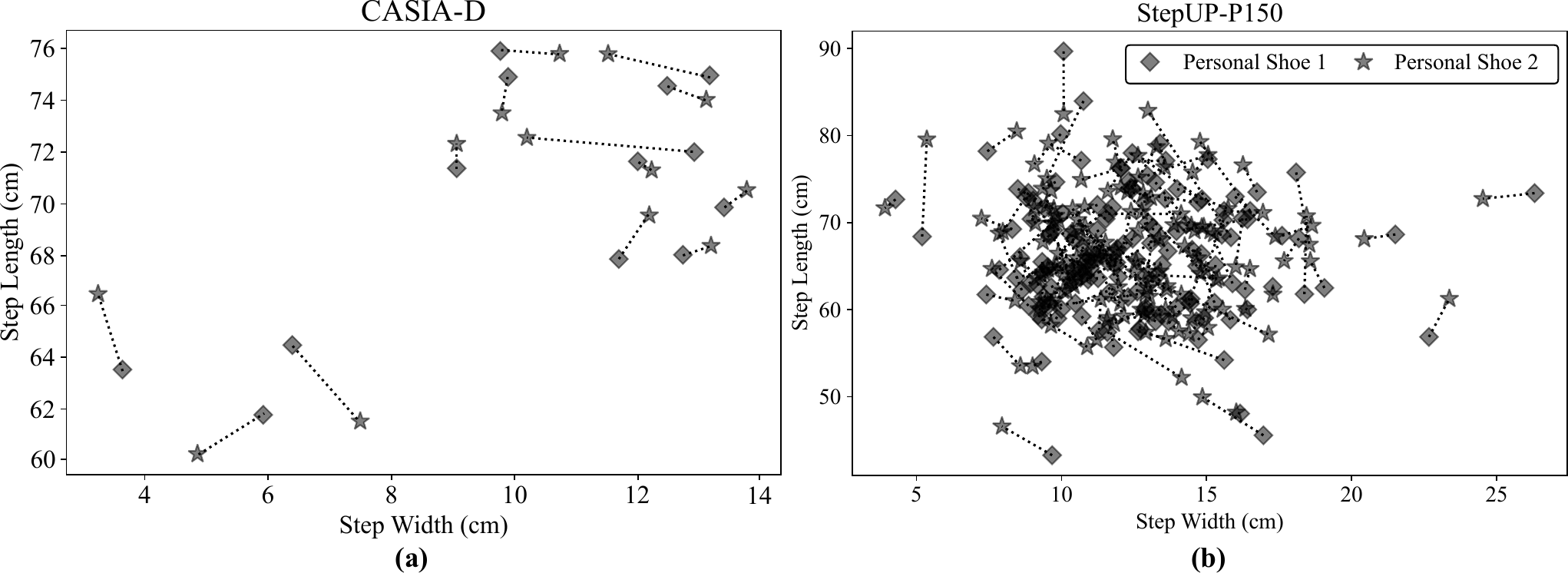}
    \caption{Average step length and step width for each participant while wearing two personal pairs of shoes from the (a) CASIA-D shod and (b) StepUP-P150 databases. Note: Dotted lines connect markers associated with the same participant, with the two shoe types represented by diamond and star markers, respectively.}
    \label{fig:spatiotemporal_scatter}
\end{figure}

Fig.~{\ref{fig:spatiotemporal_scatter}} further illustrates this variability, showing average step length and step width (also called support base) for the 13 participants in the CASIA-D shod dataset and 150 participants in the StepUP-P150 dataset, each wearing two distinct pairs of personal shoes. 
A prior study using the CASIA-D dataset {\cite{Connor2015}} found these two gait parameters to be among the most identifying features for recognizing users across changes in footwear, contributing to a classification model that achieved 90.5\% identification accuracy with a single step and 99.5\% with five steps.
However, while these two parameters may be sufficient to distinguish the 13 CASIA-D participants with near-perfect separability, StepUP-P150 contains many individuals with overlapping values, highlighting the increased challenge and realism of person recognition in larger and more diverse populations.

%%%%%%%%%%%%%%%%%%%%%%%%%%%%%%%%%%%%%%%%%%%%%%%%%%%%%%%%%%%%%%%%%%%%%%%%%%%%%%%%%%%
%%%%%%%%%%%%%%%%%%%%%%%%%%%%%%%%%%%%%%%%%%%%%%%%%%%%%%%%%%%%%%%%%%%%%%%%%%%%%%%%%%%

%%%%%%%%%%%%%%%%%%%%%%%%%%%%%%%%%%%%%%%%%%%%%%%%%%%%%%%%%%%%%%%%%%%%%%%%%%%%%%%%%%%
%%%%%%%%%%%%%%%%%%%%%%%%%%%%%%%%%%%%%%%%%%%%%%%%%%%%%%%%%%%%%%%%%%%%%%%%%%%%%%%%%%%
%%%%%%%%%%%%%%%%%%%%%%%%%%%%%%%%%%%%%%%%%%%%%%%%%%%%%%%%%%%%%%%%%%%%%%%%%%%%%%%%%%%
\section*{Usage Notes}

% The Usage Notes should contain brief instructions to assist other researchers with reuse of the data. This may include discussion of software packages that are suitable for analysing the assay data files, suggested downstream processing steps (e.g. normalization, etc.), or tips for integrating or comparing the data records with other datasets. Authors are encouraged to provide code, programs or data-processing workflows if they may help others understand or use the data. Please see our code availability policy for advice on supplying custom code alongside Data Descriptor manuscripts.

% For studies involving privacy or safety controls on public access to the data, this section should describe in detail these controls, including how authors can apply to access the data, what criteria will be used to determine who may access the data, and any limitations on data use. 

% Brief instructions to assist other researchers with reuse of the data
% Examples

% (1) Discussion of software packages that are suitable for analysing the assay data files
Footstep data are provided in NPZ (.npz) and MAT (.mat) formats, while supplementary metadata is stored in CSV files. 
These formats are readily compatible with standard Python and MATLAB toolboxes, facilitating easy import and analysis.
The size of the Python-compatible (NPZ) dataset is approximately 50 GB, while the MATLAB-compatible (MAT) dataset is approximately 118 GB.
The complete collection of UNB StepUP-P150 files can be accessed on \emph{FRDR}{\cite{UNB-StepUP}}(\url{https://doi.org/10.20383/103.01285}), along with scripts for importing and working with the data in both Python and MATLAB (see Code Availability).
% (2) Suggested downstream processing steps (e.g. normalization, etc.)
Additional Python scripts are provided to support further processing, including code for data normalization as described in the Footstep Normalization section. 
This script offers several normalization options (see Table~{\ref{tab:preprocessing}}), enabling users to apply their preferred downstream processing methods. 
While the dataset includes preprocessed outputs from two distinct pipelines to support quick prototyping and analysis, the script offers flexibility to explore a wide range of alternative preprocessing configurations.
% (3) Tips for integrating or comparing the data records with other datasets
Finally, scripts are also provided to extract several standard gait features (e.g., peak pressure images, COP and GRF time series, and spatiotemporal gait parameters like step length and width) presented in the Technical Validation section.
These are supplied to assist in benchmarking and in integrating or comparing the current dataset with other datasets.

%%%%%%%%%%%%%%%%%%%%%%%%%%%%%%%%%%%%%%%%%%%%%%%%%%%%%%%%%%%%%%%%%%%%%%%%%%%%%%%%%%%
%%%%%%%%%%%%%%%%%%%%%%%%%%%%%%%%%%%%%%%%%%%%%%%%%%%%%%%%%%%%%%%%%%%%%%%%%%%%%%%%%%%
%%%%%%%%%%%%%%%%%%%%%%%%%%%%%%%%%%%%%%%%%%%%%%%%%%%%%%%%%%%%%%%%%%%%%%%%%%%%%%%%%%%
\section*{Code Availability}

% For all studies using custom code in the generation or processing of datasets, a statement must be included under the heading "Code availability", indicating whether and how the code can be accessed, including any restrictions to access. This section should also include information on the versions of any software used, if relevant, and any specific variables or parameters used to generate, test, or process the current dataset. 

Custom scripts designed for processing and technical validation are provided in the data repository as well as on the dataset's companion GitHub page (\url{https://github.com/UNB-StepUP/StepUP-P150}) to support ongoing improvements and optimizations. 
These scripts were created using MATLAB (The MathWorks, Inc., Natick, Massachusetts, United States, 2023a) and Python (Python Software Foundation, 3.11). 
The Python scripts require specific libraries, which are listed in the requirements.txt file. 
This file enables library installation through The Python Package Index (PyPI, https://pypi.org) or the Anaconda software distribution (2024.02, https://www.anaconda.com). 
Instructions regarding the use and execution of the custom code are provided in the accompanying README file.

%%%%%%%%%%%%%%%%%%%%%%%%%%%%%%%%%%%%%%%%%%%%%%%%%%%%%%%%%%%%%%%%%%%%%%%%%%%%%%%%%%%
%%%%%%%%%%%%%%%%%%%%%%%%%%%%%%%%%%%%%%%%%%%%%%%%%%%%%%%%%%%%%%%%%%%%%%%%%%%%%%%%%%%
%%%%%%%%%%%%%%%%%%%%%%%%%%%%%%%%%%%%%%%%%%%%%%%%%%%%%%%%%%%%%%%%%%%%%%%%%%%%%%%%%%%
\bibliography{ref}

% For journal articles, DOIs should be included for works in press that do not yet have volume or page numbers. For other journal articles, DOIs should be included uniformly for all articles or not at all. We recommend that you encode all DOIs in your bibtex database as full URLs, e.g. https://doi.org/10.1007/s12110-009-9068-2.

%%%%%%%%%%%%%%%%%%%%%%%%%%%%%%%%%%%%%%%%%%%%%%%%%%%%%%%%%%%%%%%%%%%%%%%%%%%%%%%%%%%
%%%%%%%%%%%%%%%%%%%%%%%%%%%%%%%%%%%%%%%%%%%%%%%%%%%%%%%%%%%%%%%%%%%%%%%%%%%%%%%%%%%
%%%%%%%%%%%%%%%%%%%%%%%%%%%%%%%%%%%%%%%%%%%%%%%%%%%%%%%%%%%%%%%%%%%%%%%%%%%%%%%%%%%
\section*{Acknowledgements} 

 In addition to the authors, many others contributed meaningfully to this dataset. We would like to thank those who helped with the collection of the data, including Erica Cluff, Erin Kierstead, Ryan Sullivan, Sarah Boyd, and Morva Mohammedzadeh Dogaheh. We would also like to thank those who contributed, in part, to the data curation of the dataset, including Sarah Boyd, Morva Mohammedzadeh Dogaheh, Chitom Nsofor, and Grace Sanford. Finally, we thank the funding and project partners who made this project possible, including CyberNB, Knowledge Park, Stepscan Technologies, the New Brunswick Innovation Foundation, the Atlantic Canada Opportunities Agency, and the Natural Sciences and Engineering Research Council of Canada (NSERC) Alliance grants program. 

%%%%%%%%%%%%%%%%%%%%%%%%%%%%%%%%%%%%%%%%%%%%%%%%%%%%%%%%%%%%%%%%%%%%%%%%%%%%%%%%%%%
%%%%%%%%%%%%%%%%%%%%%%%%%%%%%%%%%%%%%%%%%%%%%%%%%%%%%%%%%%%%%%%%%%%%%%%%%%%%%%%%%%%
%%%%%%%%%%%%%%%%%%%%%%%%%%%%%%%%%%%%%%%%%%%%%%%%%%%%%%%%%%%%%%%%%%%%%%%%%%%%%%%%%%%
\section*{Author contributions statement}

Following the CRediT (Contributor Roles Taxonomy), the author contributions are as follows: 

\textbf{RL}: Conceptualization, Methodology, Software, Investigation, Formal Analysis, Data Curation, Writing - Original Draft, Writing - Review \& Editing, Visualization. 

\textbf{AP}: Conceptualization, Methodology, Software, Formal Analysis, Writing - Original Draft, Writing - Review \& Editing, Visualization, Supervision. 

\textbf{AS}: Investigation, Writing - Review \& Editing.

\textbf{EM}: Methodology, Investigation, Data Curation, Writing - Review \& Editing. 

\textbf{SK}: Investigation, Writing - Review \& Editing.

\textbf{SB}: Investigation, Writing - Review \& Editing.

\textbf{AT}: Software, Formal Analysis, Data Curation, Writing - Original Draft, Writing - Review \& Editing, Visualization. 

\textbf{ES}: Conceptualization, Methodology, Formal Analysis, Data Curation, Writing – Original Draft, Writing - Review \& Editing, Visualization, Resources, Supervision, Project Administration, Funding Acquisition.

%Must include all authors, identified by initials, for example:
%A.A. conceived the experiment(s), A.A. and B.A. conducted the experiment(s), C.A. and D.A. analysed the results. All authors reviewed the manuscript. 

\section*{Competing interests} %(mandatory statement)

The authors declare no competing interests.

%The corresponding author is responsible for providing a \href{https://www.nature.com/sdata/policies/editorial-and-publishing-policies#competing}{competing interests statement} on behalf of all authors of the paper. This statement must be included in the submitted article file.

\end{document}